%% file: 00_main.tex
\documentclass[sw]{iosart2x_arxiv}

\usepackage[table,xcdraw]{xcolor}
\usepackage[english]{babel}
\usepackage[utf8x]{inputenc}
\usepackage{amsmath}
\usepackage{balance}
\usepackage{todonotes}
\usepackage{enumitem}
\usepackage{booktabs}
\usepackage{multirow}
\usepackage{makecell}
\usepackage[table,xcdraw]{xcolor}
\usepackage[flushleft]{threeparttable}
\usepackage{pgfplots}
\usepackage[export]{adjustbox} 
\usepackage{subfig}
\usepackage{listings}
\pgfplotsset{compat=1.5} 

\usepackage{pifont}
\usepackage[english]{babel}
\usepackage[utf8x]{inputenc}
\usepackage{amsmath}
\usepackage{graphicx}
\usepackage{todonotes}
\usepackage{url}
\usepackage{diagbox}
\usepackage{soul} 
\usepackage{bm}
\usepackage{multirow}
\usepackage{colortbl}
\usepackage{fancyvrb} 
\usepackage{soul} 
\usetikzlibrary{pgfplots.groupplots}

\newcommand{\eat}[1]{}
\newtheorem{definition}{Definition}
\newcommand\schema[1]{{\normalfont\fontfamily{cmvtt}\selectfont #1}}

\newcommand{\sg}[1]{#1}
\newcommand{\ed}[1]{#1}

\newcommand{\sgg}[1]{#1}

\definecolor{lightgreen}{rgb}{0.6, 1, 0.6}
\DeclareRobustCommand\revA[1]{#1}
\DeclareRobustCommand\ed[1]{#1}

\firstpage{1}
\lastpage{1}

\begin{document}

\begin{frontmatter}

\title{EventKG - the Hub of Event Knowledge on the Web 
- and Biographical Timeline Generation
}

\runningtitle{EventKG - the Hub of Event Knowledge on the Web 
- and Biographical Timeline Generation}

\author[A]{\inits{S.G.}\fnms{Simon} \snm{Gottschalk}\ead[label=e1]{gottschalk@L3S.de}%
} and
\author[A]{\inits{E.D.}\fnms{Elena} \snm{Demidova}\ead[label=e2]{demidova@L3S.de}
\thanks{Corresponding author. \printead{e2}.}
}
\runningauthor{S. Gottschalk and E. Demidova}
\address[A]{L3S Research Center, \orgname{Leibniz Universit\"at Hannover},
\cny{Hannover, Germany}\printead[presep={\\}]{e1,e2}}

\begin{abstract}
	One of the key requirements to facilitate the semantic analytics of  
	information regarding contemporary and historical events 
	on the Web, in the news and in social media is the availability of reference knowledge repositories containing comprehensive representations of events, entities and temporal relations.
    Existing knowledge graphs, with popular examples including DBpedia, YAGO and Wikidata, focus mostly on entity-centric information and are insufficient in terms of their coverage and completeness with respect to events and temporal relations.
    In this article we address this limitation, 
    formalise the concept of a temporal knowledge graph
    and present its instantiation - EventKG. 
	EventKG is a multilingual event-centric temporal knowledge graph that incorporates over 690 thousand events and over 2.3 million temporal relations obtained from several large-scale knowledge graphs and semi-structured sources and makes them available through a canonical RDF representation.
Whereas popular entities often possess hundreds of relations within a temporal knowledge graph such as EventKG, generating a concise overview of the most important temporal relations for a given entity is a challenging task.
In this article we demonstrate an application of EventKG to 
    biographical timeline generation, where we adopt a distant supervision method to identify relations most relevant for an entity biography. 
Our evaluation results provide insights on the characteristics of EventKG 
and demonstrate the effectiveness of the proposed biographical timeline generation method.
\end{abstract}


\end{frontmatter}

\input{01_introduction}


\input{02a_motivation}



\input{03_definitions}


\input{04a_model}

\input{04b_ekg_tkg}

\input{04c_extraction}

\input{04d_running_example}



\input{06a_characteristics}

\input{06b_ekg_evaluation}
\input{06c_ekg_v2}


\input{05a_method}
\input{05b_running_example}


\input{07a_dataset}

\input{07b_evaluation}

\input{08_related}

\input{09_conclusion}


\newpage
\subsubsection*{Acknowledgements} This work was partially funded by the
EU Horizon 2020 under ERC grant ``ALEXANDRIA'' (339233) and MSCA-ITN-2018 
``Cleopatra'' (812997), the Federal Ministry of Education and Research (BMBF) under ``Data4UrbanMobility'' (02K15A040) and ``Simple-ML'' (01IS18054).

\bibliographystyle{ios1}           
\bibliography{00_main}        

\input{10_appendix}

\end{document}

%% file: 01_introduction.tex
\section{Introduction}
\label{sec:intro}

\textit{Motivation:}
The amount of event-centric information regarding contemporary and historical events of global importance, such as the US elections, the 2018 Winter Olympics and the Syrian Civil War, constantly grows on the Web, in the news sources and within social media. 
\ed{
In the literature, an event is typically described as something that happens at a specific time and location \cite{AllanPL17}. 
Events considered in this work are real-world happenings of societal importance, with examples including military conflicts, sports tournaments and political elections. 
In particular, we consider events, entities they involve and temporal relations - i.e. real-world relations between events and entities valid over a time period. 
}

Efficiently accessing and analysing large-scale event-centric and temporal information is crucial for a variety of real-world applications in the fields of Semantic Web, NLP and Digital Humanities. 
In Semantic Web and NLP, these applications include timeline generation \cite{Althoff:2015, gottschalk2018demo} and Question Answering \cite{HoffnerWMULN17, Huang:2019}.
In Digital Humanities, multilingual event repositories can facilitate cross-cultural studies analysing language-specific and community-specific views on historical and contemporary events (examples of such studies can be seen in \cite{GottschalkDBR17, Rogers:2013}). 
Furthermore, event-centric knowledge graphs can facilitate the reconstruction of histories as well as networks of people and organisations over time \cite{ROSPOCHER2016132, AlBadrashiny:2017}.
One of the pivotal pre-requisites to facilitate effective analytics of events is the availability of knowledge repositories providing reference information regarding events, involved entities and their temporal relations. 

\textit{Limitations of the existing sources of event-centric and temporal information:}
Currently, event representations and temporal relations are spread across heterogeneous sources. 
First, large-scale knowledge graphs (KGs) (i.e. graph-based knowledge repositories \cite{Faerber:2016} such as Wikidata \cite{Erxleben:2014}, DBpedia \cite{dbpedia-swj} and YAGO \cite{Mahdisoltani:2014}) typically focus on entity-centric knowledge. Event-centric information included in these sources is often not clearly identified as such, can be incomplete and is mostly restricted to named events and encyclopaedic knowledge.

\ed{
For example, as discussed later in Section \ref{sec:characteristics}, out of $322,669$ events included in EventKG V1.1, only $18.7\%$ are classified using the \schema{dbo:Event} class in the English DBpedia as of 12/2017. Furthermore, event descriptions in the existing knowledge graphs often lack the key properties, i.e. times and locations. For example, according to our analysis, only $33\%$ of events in Wikidata provided temporal and $11.7\%$ spatial information at that time.
}

Second, a variety of manually curated semi-structured sources (e.g. the Wikipedia Current Events Portal (WCEP) \cite{Tran:2014} and multilingual Wikipedia event lists) contain information on contemporary events. However, the lack of structured representations of events and temporal relations in these sources hinders their direct use in real-world applications, e.g. through semantic technologies. 
Overall, a comprehensive integrated view on contemporary and historical events and their temporal relations is still missing. EventKG will help to overcome these limitations.

\ed{
An additional source of event-centric information on the Web are the recently proposed knowledge graphs containing events obtained from unstructured news sources using Information Extraction methods (such as \cite{ROSPOCHER2016132, Yuan:2018, Prasojo:2018, Leetaru:2013, Boschee:2015}). 
These knowledge graphs are potentially highly noisy (e.g. \cite{ROSPOCHER2016132} reports an extraction accuracy of only $0.55$). 
Due to significant differences in quality and event granularity, the integration of events from these sources with the information in the established knowledge repositories such as DBpedia or Wikidata within a common knowledge graph does not appear meaningful. 
These event sources as well as the corresponding Information Extraction methods for unstructured news articles are out of scope of this work. 
}

\textit{A temporal knowledge graph \& EventKG:}
In this article we formalise the concept of a temporal knowledge graph
that interconnects real-world entities and events using temporal relations valid over a time period.
Furthermore, we present an instantiation of a temporal knowledge graph - EventKG.
EventKG takes an important step to facilitate a global view on events and temporal relations currently spread across entity-centric knowledge graphs and manually curated
semi-structured sources. 
EventKG integrates this knowledge in an efficient light-weight fashion, enriches 
it with additional features such as indications of relation strengths and event popularity, adds provenance information 
and makes all this information available through a canonical RDF representation. 
Through the light-weight integration and fusion of event-centric and temporal information from different sources, EventKG enables to increase coverage and completeness of this information. For example, EventKG increases the coverage of locations and dates for Wikidata events it contains by $14.43\%$ and $17.82\%$, correspondingly (see Table \ref{tab:named_events_comparison} in Section \ref{sec:characteristics} for more detail).
Furthermore, relation strengths and event popularity provided by EventKG are the characteristics that gain the key relevance given the rapidly increasing amount of event-centric and temporal data on the Web and the resulting information overload. 

EventKG was first introduced in \cite{Gottschalk:2018}.
Compared to \cite{Gottschalk:2018}, in this article we formally 
introduce the concept of a temporal knowledge graph, provide 
more details on the algorithms adopted for the EventKG generation and the corresponding evaluation results. Furthermore, we 
present a method that facilitates an application 
of EventKG to biographical timeline generation.
\sg{We make EventKG, including the dataset, a SPARQL endpoint, the code and evaluation data, as well as the benchmarks created for the biographical timeline generation available online\footnote{\url{http://eventkg.l3s.uni-hannover.de/}}.}

\begin{figure*}[t]
  \includegraphics[width=\linewidth]{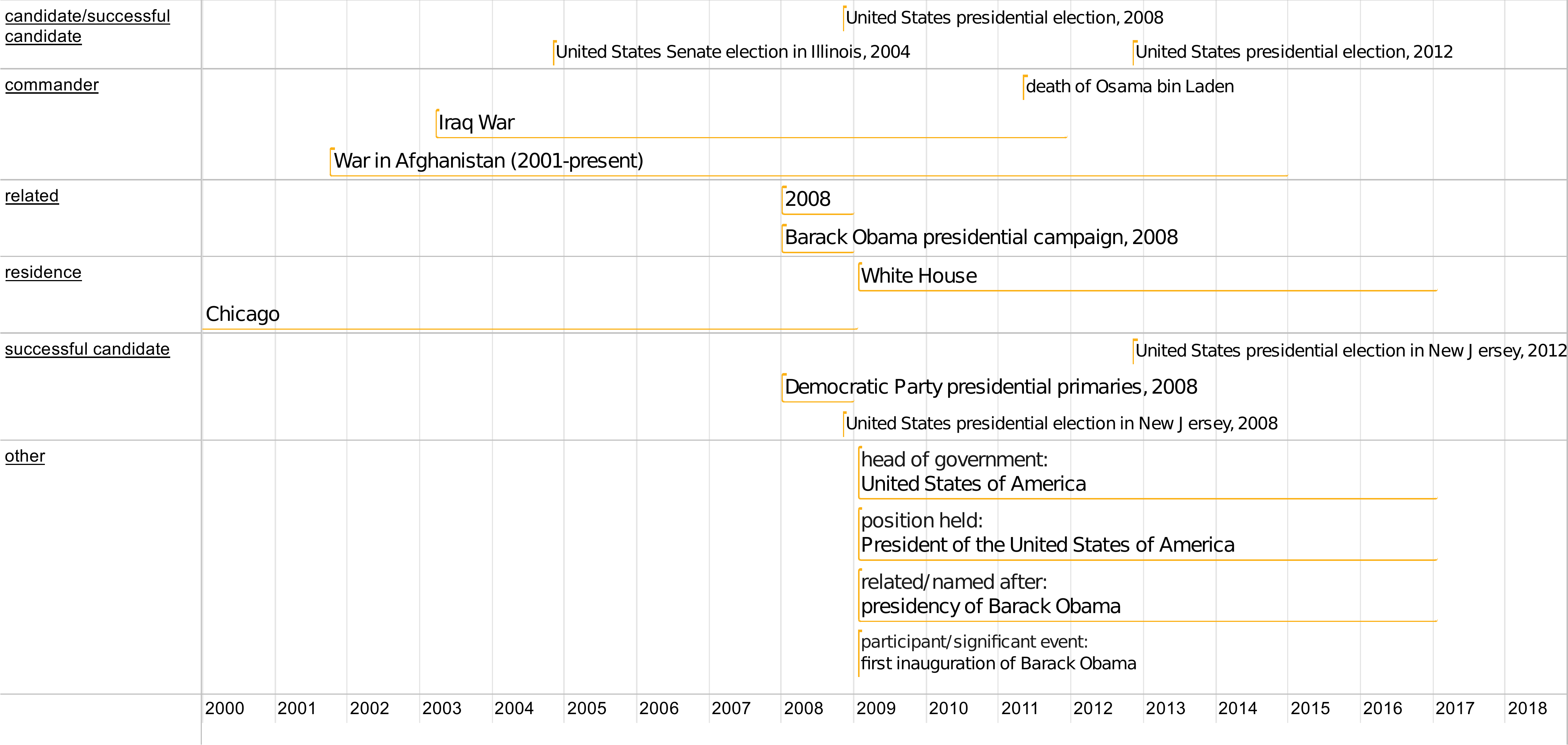}
  \caption{
  \ed{An excerpt of the biographical timeline for the entity Barack Obama, 
  generated from the EventKG knowledge graph using a proposed model trained on the Wikipedia abstracts of other entities (BS-ENC benchmark). 
  Orange lines represent the temporal validity of the relations. 
  Each row corresponds to a predicate characterising the relation (e.g. \textit{commander}) to the specific event or entity (e.g. \textit{Iraq War}).}
  }
  \label{fig:timeline_example}
\end{figure*}


\textit{Generation of Biographical Timelines using a Temporal Knowledge Graph:}
A popular entity such as an influential person, a city or a large organisation can impose hundreds of temporal relations within a temporal knowledge graph. 
For example, the entity Barack Obama possesses $2,608$ temporal relations in EventKG. 
Identifying the most important temporal relations
within the temporal knowledge graph to provide a concise overview for a given entity becomes a challenging task in these settings.

Timelines are an effective method to provide a visual overview of entity-centric temporal information, such as temporal relations in a knowledge graph \cite{Althoff:2015}. 
In particular, biographical timelines describe significant happenings in a person's life and typically include events of major relevance from the personal perspective such as birth, education and career. 
Figure \ref{fig:timeline_example} illustrates a biographical timeline for Barack Obama, which includes places where Barack Obama lived (first Chicago and then the White House), important events he was involved in (e.g. the Iraq War) and the major political positions he held (e.g. the President of the United States). This timeline also indicates the temporal validity of these relations.

In this article we present an approach for the generation of biographical timelines from a temporal knowledge graph.
To generate such timelines, we propose a distant supervision method, where we train the relevance model using external sources containing biographical and encyclopaedic texts. With that model, we extract the most relevant biographical data from the temporal knowledge graph concisely describing a person's life, while using features such as relation strength and event popularity information contained in EventKG, as well as predicate labels.
The results of our user evaluation demonstrate that this approach is able to generate high quality biographical timelines while significantly outperforming a state-of-the-art baseline for timeline generation: our timelines were preferred over the baseline's timelines in approximately $68\%$ of the cases. 

Overall, our contributions in this article are as follows:

\begin{enumerate}
\item [1] We formally define the concept of a temporal knowledge graph $TKG$ that incorporates entities, events and temporal relations.
\item [2] We present an instantiation of $TKG$: EventKG - a multilingual RDF
 knowledge graph that incorporates over $690$ thousand events and over $2.3$ million temporal relations in V1.1. 
 We provide insights on the extraction and fusion methods adopted to generate the EventKG knowledge graph and their quality. 
\item [3] We define the problem of biographical timeline generation from a temporal knowledge graph and present our method based on distant supervision.
\item [4] We demonstrate the effectiveness of the proposed timeline generation method in a user study.
\end{enumerate}

The remainder of this article is organised as follows: 
First, in Section \ref{sec:relevance} we motivate the need for a temporal knowledge graph and introduce a running example.
In Section \ref{sec:tkg}, we formally define the concepts of a temporal knowledge graph and a biographical timeline. Then, in Section \ref{sec:eventkg}, we describe EventKG, including its 
RDF data model and the extraction pipeline.
In Section \ref{sec:char-eval}, we provide statistics and evaluation results of the data contained in EventKG. 
Our approach towards biographical timeline generation using temporal knowledge graph is presented in Section \ref{sec:timelines}. 
The experimental setup and evaluation of the biographical timelines generated with our approach using EventKG is provided in Section \ref{sec:eval-timeline}. 
Related work is discussed in Section \ref{sec:related}. Finally, we discuss our findings and provide a conclusion in Section \ref{sec:conclusion}.

%% file: 02a_motivation.tex
\section{Motivation}
\label{sec:relevance}

\begin{table*}[!t]
      \centering
    \caption{All events connected with Barack Obama in EventKG that started between November 4 and November 16, 2011.}
          \label{tab:wwii_timeline}

      \footnotesize
      \centering
\begin{tabular}{|r||p{4cm}|p{7.5cm}|}
\hline
\multicolumn{1}{|c||}{\textbf{\makecell{Start Date}}} & \multicolumn{1}{c|}{\textbf{Sources}} & \multicolumn{1}{c|}{\textbf{Description}}                                                                                                                                                                       \\ \hline \hline
Nov 4 & YAGO, Wikidata, DBpedia\textsubscript{EN}, DBpedia\textsubscript{FR}, DBpedia\textsubscript{RU} & 2011 G20 Cannes summit                                                                                                                                        \\ \hline
Nov 11 & YAGO, Wikidata, DBpedia\textsubscript{EN} & 2011 White House shooting \\ \hline 
Nov 16 & Wikipedia\textsubscript{EN} & The President of the United States Barack Obama visits Australia to commemorate the 60th anniversary of the ANZUS alliance. \\ \hline
\end{tabular}
  
\end{table*}

\begin{table*}[t]
\centering
\footnotesize
    \caption{Most linked events in the English (EN) and the Russian (RU) Wikipedia.}
    \label{tab:top_3_events}
\begin{tabular}{|l||l|r||l|r|}
\hline
\textbf{Rank} & \multicolumn{1}{c|}{\textbf{Event (EN)}} & \multicolumn{1}{c||}{\textbf{\#Links (EN)}} & \multicolumn{1}{c|}{\textbf{Event (RU)}} & \multicolumn{1}{c|}{\textbf{\#Links (RU)}} \\ \hline \hline
1 & World War II & 189,716 & World War II & 25,295 \\ \hline
2 & World War I & 99,079 & World War I & 22,038 \\ \hline
3 & American Civil War & 37,672 & October Revolution & 7,533 \\ \hline
4 & FA Cup & 20,640 & Russian Civil War & 7,093 \\ \hline
\end{tabular}
\end{table*}

\begin{table*}[t]
\centering
\footnotesize
\caption{Top-4 persons mentioned jointly with the financial crisis (2007–2008) per language.}
\label{tab:crisis_relation_strength}
\begin{tabular}{|l||l|l|l|l|l|}
\hline
\textbf{Rank} & \multicolumn{1}{c|}{\textbf{EN}} &  \multicolumn{1}{c|}{\textbf{FR}} & \multicolumn{1}{c|}{\textbf{DE}} & \multicolumn{1}{c|}{\textbf{RU}} & \multicolumn{1}{c|}{\textbf{PT}} \\ \hline \hline
\multicolumn{1}{|l||}{\textbf{1}} & Barack Obama & Kevin Rudd & Barack Obama & Michael Moore & Barack Obama \\ \hline
\multicolumn{1}{|l||}{\textbf{2}} & George W. Bush & John Howard & Geir Haarde & Roman Abramovich & José Sócrates \\ \hline
\multicolumn{1}{|l||}{\textbf{3}} & Joseph Stiglitz & Don Cheadle & George W. Bush & Adam McKay & Pope Benedict XVI \\ \hline
\multicolumn{1}{|l||}{\textbf{4}} & Ben Bernanke & Ben Bernanke & Wolfgang Schäuble & Mikhail Prokhorov & Gordon Brown \\ \hline
\end{tabular}
\end{table*}

Our society faces an unprecedented number of events that impact multiple communities across language and community borders. In this context, the efficient access to event-centric multilingual information originating from different sources, as facilitated by EventKG, is of utmost importance for several scientific communities, including Semantic Web, NLP and Digital Humanities and a variety of applications, including 
timeline generation, question answering, as well as cross-cultural and cross-lingual event-centric analytics. 

Timeline generation is an active research area \cite{Althoff:2015, gottschalk2018demo}, where the focus is to generate a timeline (i.e. a chronologically ordered selection) of events and temporal relations for entities from a knowledge graph.
In this article we focus on the application of EventKG to the automated generation of timelines representing people biographies.
In this task, information regarding event popularity and relation strength available in EventKG in a combination with a benchmark extracted from external biographical sources can enable the selection of the most relevant timeline entries.

EventKG facilitates the generation of detailed timelines containing complementary information originating from different reference sources, potentially resulting in more complete timelines and event representations. For example, Table \ref{tab:wwii_timeline}
illustrates an excerpt from the timeline for the query \textit{``What were the events related to Barack Obama between November 4 and November 16, 2011?''} generated using EventKG. 
The last event in the timeline in Table \ref{tab:wwii_timeline} about Obama visiting Australia extracted from an English Wikipedia event list (``2011 in Australia''\footnote{\url{https://en.wikipedia.org/wiki/2011_in_Australia}}) is not contained in any of the reference knowledge graphs used to populate EventKG (Wikidata, DBpedia, and YAGO). 
The reference sources of the other two events include complementary information. For example, while the ``2011 White House shooting'' is assigned a start date in Wikidata, it is not connected to Barack Obama in that source.

\ed{
An important application of EventKG is cross-cultural and cross-lingual 
analytics. Such analytics can provide insights on the differences in the 
event perception and interpretation across communities.
For example, event popularity and relation strength between events and entities varies across different cultural and linguistic contexts. 
These differences can be observed and analysed using information provided by EventKG.}
For example, Table \ref{tab:top_3_events} presents the top-4 most popular events in the English vs. the Russian Wikipedia language editions as measured by how often these events are referred, i.e. linked to in the respective Wikipedia language edition. Whereas both Wikipedia language editions mention events of global importance, here the two World Wars, most frequently, the other most popular events (e.g. ``October Revolution'' and ``American Civil War'') are language-specific.
The relation strength between events and entities in specific language contexts can be inferred by counting their joint mentions in Wikipedia. 
For example, Table \ref{tab:crisis_relation_strength} lists the persons most related to the financial crisis in the years 2007 and 2008 in different Wikipedia language editions. 
\ed{
This information is directly provided by EventKG.}
An EventKG application to cross-lingual timeline generation was presented in \cite{gottschalk2018demo}. In this context, EventKG-empowered interfaces can be used as a starting point to identify events controversial in their cross-cultural aspects. 
Such events can then be analysed in more detail using tools such as MultiWiki \cite{GottschalkD17} proposed in our previous work. 

\ed{
Another intended future application of EventKG is semantic event-centric question answering.}
With the provision of EventKG, it becomes possible to answer questions such as 
\textit{``Which events related to Bill Clinton happened in Washington in 1980?''} and 
\textit{``What are the most important events related to Syrian Civil War that took place in Aleppo?''} that are of interest for both cross-cultural and cross-lingual event-centric analytics (e.g. illustrated in \cite{Rogers:2013,Gottschalk:2018TPDL}) as well as question answering and semantic search applications (e.g. \cite{HoffnerWMULN17, Zheng:2017, Huang:2019, Demidova:2013QC}).

\subsection{Running Example: A Biographical Timeline of Barack Obama}

\sg{As a running example throughout this article, we will use the task of biographical timeline generation for the entity Barack Obama. First, we will illustrate the heterogeneity of data about Barack Obama available in 
the reference knowledge graphs used to populate EventKG (Wikidata, DBpedia, YAGO and Wikipedia), 
and the extraction and integration of this data into a canonical RDF representation in EventKG. As mentioned above, this process leads to $2,608$ temporal relations involving Barack Obama. In order to generate a biographical timeline of Obama, 
the relevance of these relations to his biography needs to be assessed. 
We will describe the distant supervision approach and the features adopted to this task, 
which finally leads to the timeline depicted in Figure~\ref{fig:timeline_example}.}

%% file: 03_definitions.tex
\section{A Temporal Knowledge Graph and Biographical Timelines}
\label{sec:tkg}

A temporal knowledge graph $TKG$ connects real-world entities and events using temporal relations, i.e. relations valid over a time period.

\begin{definition}
\label{def:tkg}
A \emph{temporal knowledge graph} $TKG:$ $\langle$ $E_t$, $R_t$ $\rangle$ is a directed multigraph. The nodes in $E_t=E \cup \mathcal{V}$ are temporal entities, where $E$ is a set of real-world entities and $\mathcal{V}$ is a set of real-world events.
The directed edges in $R_t$ represent temporal relations of the temporal entities in $E_t$. 
\end{definition}

A temporal entity $e \in E$ represents a real-world entity such as a person, a location, an organisation or a concept. 
A temporal entity $e \in \mathcal{V}$ represents a real-world historical or contemporary event. Examples of events include cultural, sporting or political happenings. 
The temporal entities in $TKG$ are characterised through their existence time (for real-world entities) or happening time (for events).
 
\begin{definition}
A \emph{temporal entity} $e\in E_t$ represents a real-world entity or event. $e$ is annotated with a tuple $\langle$ $e_{uri}$, $e_{time}$ $\rangle$, where
$e_{uri}$ is the unique entity identifier and $e_{time}=[e_{start},e_{end}]$ denotes the existence time of the entity (for $e \in E$) or the happening time of the event (for 
$e \in \mathcal{V}$). 
\end{definition}
 
A temporal entity $e\in E_t$ can be assigned further properties, such as an entity type, a label and a textual description. 

A temporal relation is a binary relation of the temporal entities valid over a certain period of time. More formally:

\begin{definition}
A \emph{temporal relation} $r\in R_t$ represents a binary relation between two temporal entities. $r$ is annotated with a tuple 
$\langle r_{uri},  r_{time}, e_{i}, e_{j} \rangle$, where
$r_{uri}$ is a unique relation identifier, 
$e_{i}$ and $e_{j}$ are the temporal entities participating in the relation 
$r$ and
$r_{time}=[r_{start},r_{end}]$ denotes the validity time interval of the temporal relation. 
\end{definition}
 
The relation identifier $r_{uri}$ reflects the semantics of the 
temporal relation and is typically specified as a vocabulary term.

Given a temporal knowledge graph $TKG:\langle E_t, R_t \rangle$, 
we denote the temporal entity of user interest $e\in E_t$ for which the 
biographical timeline is generated as a \textit{timeline entity}.

A biographical timeline is a chronologically ordered list of temporal relations involving the timeline entity and relevant to that entity's biography.

\ed{
\begin{definition}
	A \emph{biographical timeline} $TL(e, bio) = (r_1,\dots,r_n)$ 
    of a timeline entity $e$
    is a chronologically ordered list of timeline entries (i.e. temporal relations involving $e$), 
    where each timeline entry $r_i$ is relevant to the entity biography $bio$.  
\end{definition}

In this article, we assume a binary notion of relevance, i.e. 
$\forall r_i \in TL(e, bio): relevance(e, r_i, bio)=1$. 

The list of timeline entries in $TL(e, bio)$ is ordered chronologically by their start time: $\forall r_i, r_j \in TL(e, bio): i \leq j \Leftrightarrow r_{i_{start}} \leq r_{j_{start}}$. 
}

An entity connected to $e$ via a timeline entry $r_i$ is denoted as a \textit{connected entity} in the following.

%% file: 04a_model.tex
\section{EventKG Knowledge Graph}
\label{sec:eventkg}

EventKG is a knowledge graph
that instantiates the temporal knowledge graph defined in Definition \ref{def:tkg},
and at the same time facilitates the integration and fusion of a variety of heterogeneous 
event representations and temporal relations extracted from several reference sources.

\sg{A \textit{reference source} is a semantic source such as a knowledge graph (e.g. Wikidata or YAGO) or a collection of articles (e.g. the French Wikipedia) used to populate EventKG.}

In the following, we present the RDF data model of EventKG in Section \ref{sec:model} and its transformation into a $TKG$ in Section \ref{sec:ekg_tkg}.
Following that we present the EventKG generation pipeline in Section \ref{sec:extraction} and illustrate the pipeline steps with our running example of Barack Obama in Section \ref{ex:data}.

\subsection{EventKG RDF Data Model}
\label{sec:model}

\textit{The goals of the EventKG RDF data model} are to facilitate a light-weight integration and fusion of heterogeneous event representations and temporal relations extracted from the reference sources, as well as to make this information available to real-world applications through an RDF representation. 
The EventKG data model is driven by the following objectives:

\begin{itemize}
\item Define the key properties of events through a canonical representation.
\item Represent temporal relations between events and entities
(including event-entity, entity-event and entity-entity relations). 
\item Include information quantifying and further describing these relations.
\item Represent relations between events (e.g. in the context of event series).
\item Support an efficient light-weight integration of event representations and temporal relations originating from heterogeneous sources.
\item Provide provenance for the information included in EventKG.
\end{itemize}

\begin{figure*}
 \centering
  \includegraphics[width=0.9\textwidth]{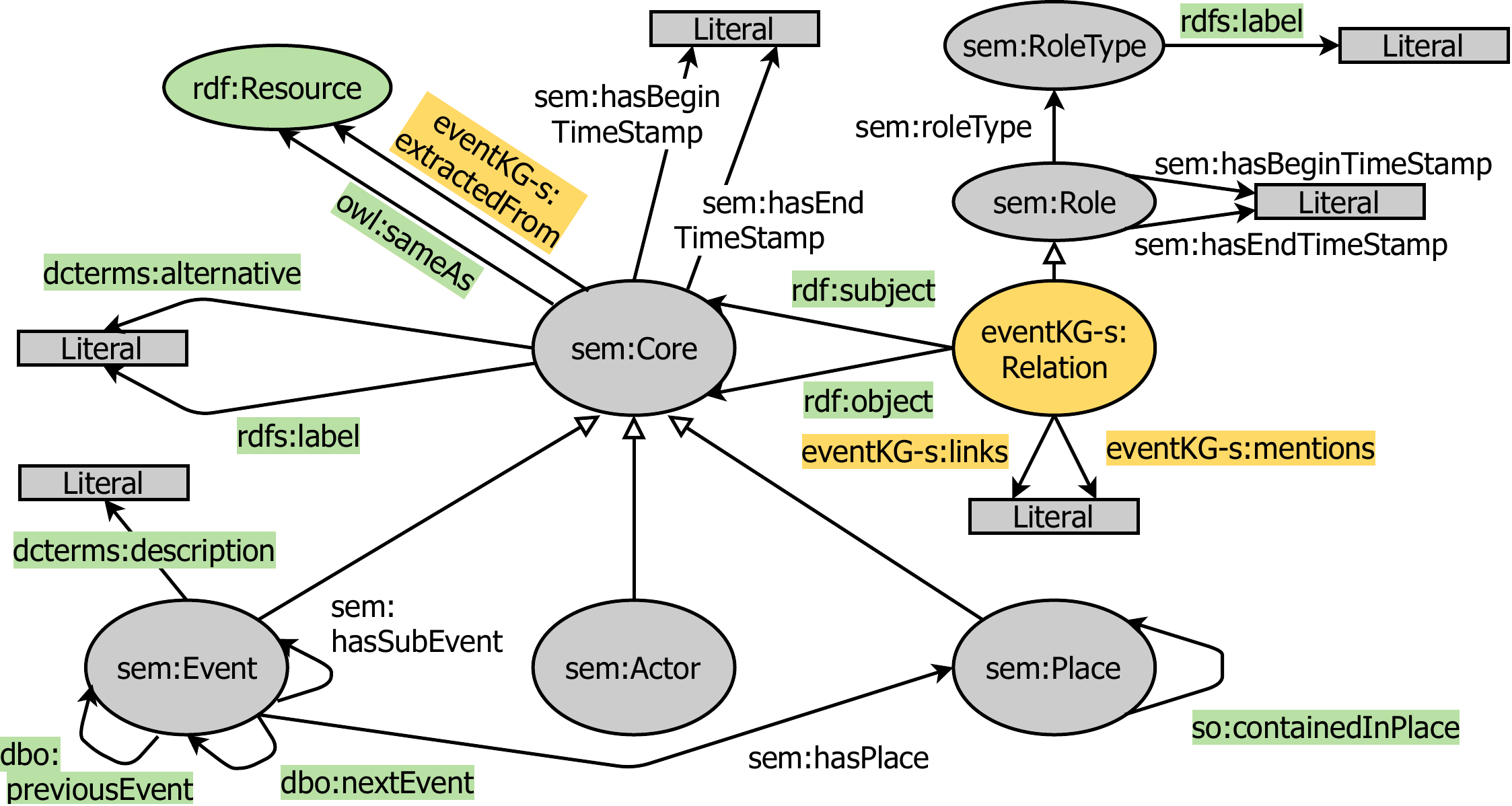}
  \caption{The EventKG schema based on SEM. Arrows with an open head denote \schema{rdfs:subClassOf} properties. Regular arrows visualize the \schema{rdfs:domain} and \schema{rdfs:range} restrictions on properties. Terms from other reused vocabularies are colored green. Classes and properties introduced in EventKG are colored orange.}
  \label{fig:schema}
\end{figure*}

\textit{EventKG schema and the Simple Event Model:} 
In EventKG, we build upon the Simple Event Model (SEM) \cite{VanHage:2011} as a basis to model events in RDF. SEM is a flexible data model that provides a generic event-centric framework. 
\ed{
The main rationale of SEM is to provide a simple model that can represent events and their key properties. Events within EventKG come from heterogeneous sources where they can be described at a different level of detail. SEM provides the lowest common denominator for event-centric information, whereas it still includes the key properties of events and their relations. The properties of events in the EventKG data model are not mandatory, such that we can also include under-specified events in EventKG, e.g. in case the corresponding temporal or geospatial information is missing in the reference sources. 
}
In addition to SEM, within the EvenKG schema, we adopt additional properties and classes to adequately represent the information extracted from the reference sources, to model temporal relations and event relations as well as to provide provenance information.
The schema of EventKG is presented in Figure \ref{fig:schema} and the used RDF namespaces are listed in Table \ref{tab:namespaces}.

EventKG is an RDF-based dataset, such that extensions to its data model are easily possible. In future work, such extensions can be performed to model 
confidence and uncertainty in the information extraction, integration and fusion, or to provide more fine-granular time information (using e.g. EDTF (Extended Date-Time Format) \cite{EDTF}). 

\sg{
\begin{table}[t]
\footnotesize
\centering
\caption{Namespaces used in the EventKG RDF model.}
\label{tab:namespaces}
\begin{tabular}{|l||r|}
\hline
\multicolumn{1}{|c||}{\textbf{Namespace prefix}} & \multicolumn{1}{c|}{\textbf{IRI}} \\ \hline \hline
\schema{so:} & http://schema.org/ \\ \hline
\schema{dbo:} & http://dbpedia.org/ontology/ \\ \hline
\schema{rdf:} & http://www.w3.org/1999/02/22-rdf-syntax-ns\# \\ \hline
\schema{rdfs:} & http://www.w3.org/2000/01/rdf-schema\# \\ \hline
\schema{dcterms:} & http://purl.org/dc/terms/rdfs: \\ \hline
\schema{sem:} & http://semanticweb.cs.vu.nl/2009/11/sem/ \\ \hline
\schema{eventKG-s:} & http://eventKG.l3s.uni-hannover.de/schema/ \\ \hline
\schema{eventKG-r:} & http://eventKG.l3s.uni-hannover.de/resource/ \\ \hline
\schema{eventKG-g:} & http://eventKG.l3s.uni-hannover.de/graph/ \\ \hline
\end{tabular}
\end{table}
}

\textit{Events and entities}: 
SEM provides a generic event representation including  topical, geospatial and temporal dimensions of an event, as well as links to its actors (i.e. entities participating in the event). Such resources are identified within the namespace \schema{eventKG-r}. 
Thus, the key classes of SEM and of the EventKG schema are \schema{sem:Event} representing events, \schema{sem:Place} representing locations and \schema{sem:Actor} representing entities participating in the events. Each of these classes is a subclass of \schema{sem:Core}, which is used to represent all entities in the temporal knowledge graph\footnote{Note that entities in EventKG are not necessarily actors in the events; temporal relations between two entities are also possible.}.
Events are connected to their locations through the \schema{sem:hasPlace} property. 
A \schema{sem:Core} instance can be assigned an existence time denoted via \schema{sem:has\-Begin\-Time\-Stamp} and \schema{sem:\-has\-End\-Time\-Stamp}. 
In addition to the SEM representation, EventKG provides textual information regarding events and entities extracted from the reference sources including labels (\schema{rdfs:label}), aliases (\schema{dcterms:\-alter\-native}) and descriptions of events (\schema{dcterms:\-description}). 

In the context of this article, the term temporal relation refers to real-world relations between events and entities valid over a period of time.
The set of temporal relations in EventKG includes event-entity, entity-event and entity-entity relations.
Temporal relations between events and entities typically connect an event and its actors (as in SEM).
A typical example of a temporal relation between two entities is a marriage. 
Temporal relations between entities can also indirectly capture information about events \cite{ROSPOCHER2016132}. For example, the DBpedia property \schema{http://dbpedia.org/property/\\acquired} can be used to represent an event of acquisition of one company by another. 
Temporal relations in SEM are limited to the situation where an actor plays a specific role in the context of an event. This yields two limitations: (i) there is no possibility to model temporal relations between events and entities where the entity acts as a subject. For example, it is not possible to directly model the fact that Barack Obama participated in the event ``Second inauguration of Barack Obama'', as the entity ``Barack Obama'' plays the subject role in this relation; and (ii) a temporal relation between two entities such as a marriage can not be modelled directly\footnote{Consider the difference between a wedding that is modelled as an event and a marriage between two people that can be modelled as a temporal relation.}. 

\begin{figure*}
 \centering
  \includegraphics[width=0.85\textwidth]{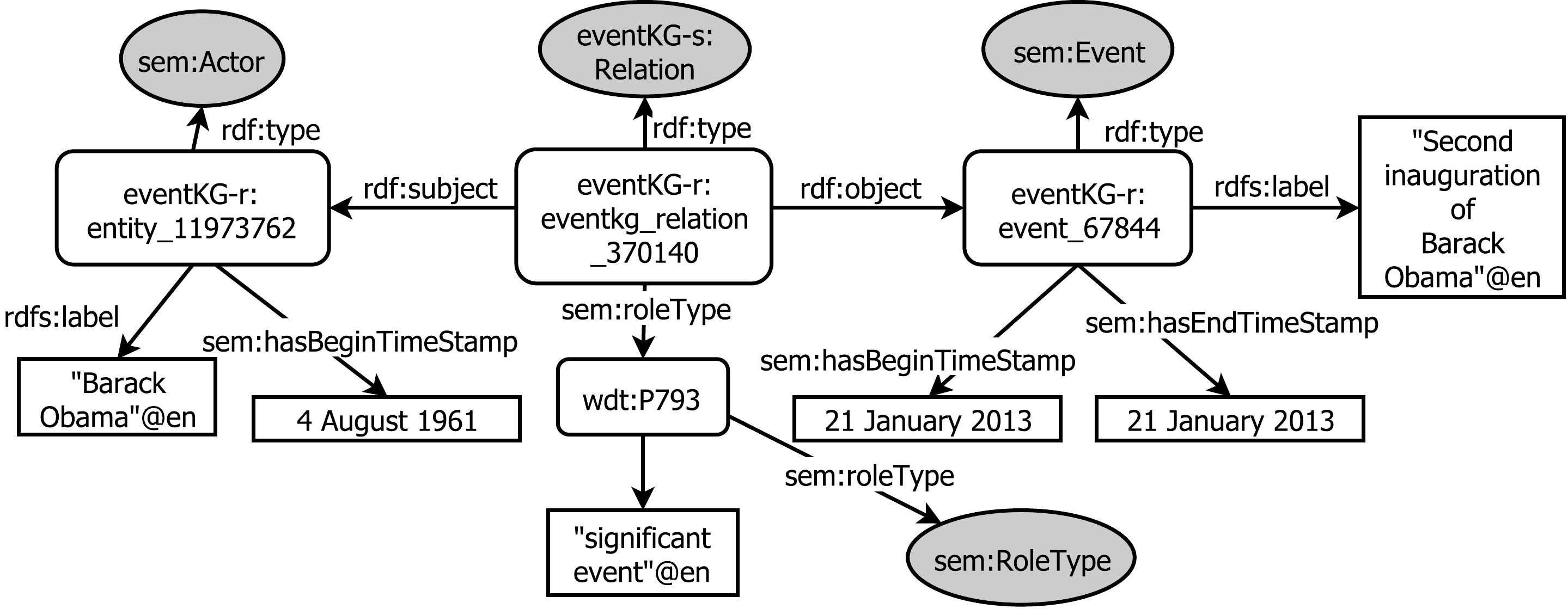}
  \caption{Example of the event representing the participation of Barack Obama in his second inauguration as a US president in 2013 as modelled in EventKG. \schema{wdt:P793} is the Wikidata identifier for the ``significant event'' property.}
  \label{fig:relation_instances}
\end{figure*}

To overcome these limitations, EventKG introduces the class \schema{eventKG-s\-:\-Re\-lation} representing relations between events and entities. 
This way of relation modelling 
facilitates flexible additional attributes describing a relation\footnote{See W3C Working Group Note from 12 April 2006
on defining N-ary Relations on the Semantic Web: \url{https://www.w3.org/TR/swbp-n-aryRelations}.}.
This class links two \schema{sem:Core} instances (each representing an event or an entity). 
The resulting relation can be annotated with a validity time and a property \schema{sem:\-Role\-Type} that characterises the relation using RDF predicates. 
Currently, the predicates are directly derived from the reference sources.
In future work, we envision the normalisation of these predicates 
by mapping them to a dedicated ontology (e.g. the DBpedia ontology\footnote{\url{https://wiki.dbpedia.org/services-resources/ontology}}).  
This way, arbitrary temporal relations between entity pairs or relations involving an entity and an event can be represented. 
This model provides flexibility to express heterogeneous temporal relations derived from the 
reference sources.
Figure \ref{fig:relation_instances} visualises the example mentioned above using the EventKG data model.

\textit{Other event and entity relations}: Relations between events (in particular sub-event, previous and next event relations)
play an important role in the context of event series (e.g. Olympic Games), 
seasons containing a number of related events (e.g. in sports), 
or events related to a certain topic (e.g. operations in a military conflict).
Sub-event relations are modelled using the \schema{so:\-has\-Sub\-Event} property. To interlink events within an event series such as the sequence of Olympic Games, the properties \schema{dbo:\-previous\-Event} and \schema{dbo:\-next\-Event} are used. 
A location hierarchy is provided through the property \schema{so:con\-tained\-InPlace}.

\textit{Towards measuring relation strength and event popularity}:
Measuring relation strength between events and entities and event popularity enables answering question like 
\ed{
\textit{``Who were the most important participants of the US Election 2016?''} or \textit{``What are the most popular events related to the Summer Olympics 2016?''}.
Relation strength and event popularity are of importance for many practical applications. For example, relation strength can help when using the knowledge graph to jointly disambiguate entities and events in text documents or in natural language questions in the context of question answering applications. 
Relation strength and event popularity can also support ranking-based applications including timeline generation and event-centric information retrieval.  
}

\ed{
Whereas the exact computation of relation strength and event popularity metrics
can be appli\-cation-depen\-dent, we include two major factors 
required for such computations,
namely \textit{links} and \textit{mentions} in the EventKG schema: 
}

\begin{itemize}[noitemsep,topsep=0pt]
\item \textit{1. Links:} This factor represents how often the description of one entity refers to another entity. Intuitively, this factor can be used to estimate the popularity of events and the strength of their relations.
In EventKG the links factor is represented through the predicate \schema{eventKG-s:\-links} in the domain of \schema{eventKG-s:\-Relation}.   
\schema{eventKG-s:\-links} denotes how often the Wikipedia article representing the relation subject links to the entity representing the  object.

\item \textit{2. Mentions:} \schema{eventKG-s:\-mentions} represents the number of relation mentions in external sources. Intuitively, this factor can be used to estimate the relation strength.
In EventKG, \schema{eventKG-s:\-mentions} denotes the number of sentences in Wikipedia that mention both, the subject and the object of the relation. 
\end{itemize}

\ed{
Links and mentions factors provided by EventKG are computed using 
sources external to the knowledge graph, such as the entire Wikipedia corpus.
Having this information included directly in the knowledge graph can help the relevant applications to obtain this information efficiently and to directly use it in 
their computations, including (but potentially not limited to) relation strength and 
event popularity metrics.
}

\textit{Provenance information}: 
EventKG provides the following provenance information: 
(i) provenance of the individual resources; 
(ii) representation of the reference sources; and 
(iii) provenance of statements.

\textit{Provenance of the individual resources:} EventKG resources typically directly correspond to the events and entities contained in the reference sources (e.g. an entity representing Barack Obama in EventKG corresponds to the DBpedia resource \schema{http://dbpedia.org/\allowbreak page/\allowbreak Barack\_Obama}). In this case, the \schema{owl:sameAs} property is used to interlink both resources. 
EventKG resources can also be extracted from a resource collection. For example, philosophy events in 2007 can be extracted from the Wikipedia event list at \schema{https://\allowbreak en.\allowbreak wikipedia.org/\allowbreak wiki/2007\_in\_philosophy}. 
In this case, the EventKG property \schema{eventKG-s:\-extracted\-From} is utilised to establish the link between the EventKG resource and the resource collection from which this resource was extracted.
Through the provenance URIs, background knowledge contained in the reference sources can be accessed.

\textit{Representation of the reference sources:} EventKG and each of the reference sources are represented through an instance of \schema{void:\-Dataset}\footnote{The VoID vocabulary \url{https://www.w3.org/TR/void/}.}. Such an instance in the namespace \schema{eventKG-g} includes specific properties of the source (e.g. its creation date as in: \schema{eventKG-g:dbpedia\_pt dcterms:created "2016-10-01"\^{}\^{}xsd:date}).


\textit{Provenance information of statements:}
A statement in EventKG is represented as a quadruple, containing a triple and a URI of the named graph it belongs to.
Through named graphs, EventKG offers an intuitive way to retrieve information extracted from the individual reference sources using SPARQL queries.

%% file: 04b_ekg_tkg.tex
\subsection{EventKG as a Temporal Knowledge Graph}
\label{sec:ekg_tkg}

A named graph such as \schema{eventKG-g:event\_kg} can be expressed as a temporal knowledge graph $TKG:$ $\langle$ $E_t$, $R_t$ $\rangle$ as follows:

\begin{itemize}
\item \textit{Entities and events}: Each instance of \schema{sem:Core} is a temporal entity $e \in E_t$ and each instance of \schema{sem:Event} is an event $v \in \mathcal{V}$, such that $E = E_t  \setminus \mathcal{V}$ is the set representing real-world entities.
\ed{
\item \textit{Time information for entities and events}: For each temporal entity $e=\langle e_{uri}, e_{time}\rangle, e \in E_t$, $e_{uri}$ is the URI of the corresponding EventKG entity. $e_{start}$ and $e_{end}$ are set according to the \schema{sem:has\-Begin\-Time\-Stamp} and \schema{sem:has\-End\-Time\-Stamp} values in the \schema{eventKG-g:event\_kg} named graph, correspondingly.
}
\ed{
\item \textit{Temporal relations with known validity times}: Each instance of \schema{eventKG-s:Relation} that has a start or an end time in the named graph is transformed into a temporal relation $r=\langle r_{uri},\allowbreak r_{time},\allowbreak e_{i},\allowbreak e_{j} \rangle \in R_t$. 
Here, $r_{uri}$ is the URI of the EventKG relation instance, 
$e_i$ is the entity connected to the \schema{eventKG-s:Relation} instance via \schema{rdf:subject}, $e_j$ is the entity connected via \schema{rdf:object} and $r_{time}$ includes the \schema{sem:has\-Begin\-Time\-Stamp} and \schema{sem:has\-End\-Time\-Stamp} relations.
}
\ed{
\item 
\textit{Indirect temporal relations}:
Information regarding the temporal validity of a relation is not always explicitly provided in EventKG. However, this information can often be derived based on the existence times of the participating entities or the happening times of the events.
For example, the validity of a ``mother'' relation can be determined using the birth date of the child entity.
We refer to such relations as \textit{indirect temporal relations}. 
Each instance of \schema{eventKG-s:Relation} that represents such an indirect temporal relation is transformed into a temporal relation $r_t = \langle r_{uri},\allowbreak r_{time},\allowbreak e_i,\allowbreak e_j\rangle \in R_t$, $r_{time} = e_{j_{time}}$.
}
\end{itemize}

%% file: 04c_extraction.tex
\subsection{EventKG Generation Pipeline}
\label{sec:extraction}
\begin{figure*}[ht]
  \centering
  \includegraphics[width=\textwidth]{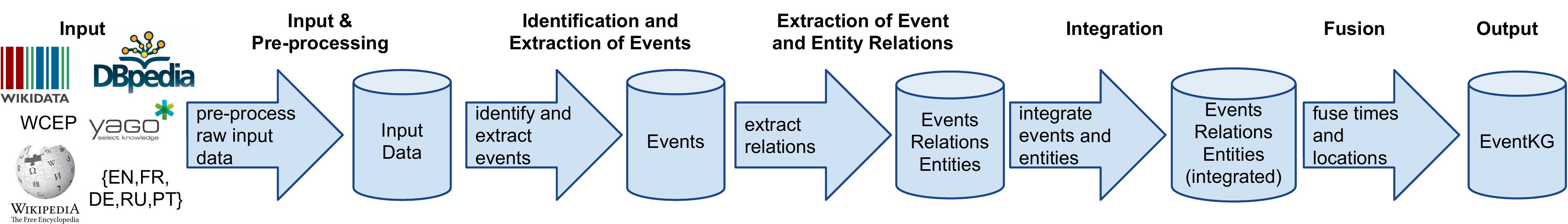}
  \caption{\sgg{The EventKG generation pipeline.}}
  \label{fig:pipeline}
\end{figure*}

The EventKG generation pipeline is shown in Figure~\ref{fig:pipeline}.

\textit{Input \& Pre-processing}: 
\ed{
First, the dumps of the reference sources in the corresponding languages are collected. 
Both Wikidata and YAGO provide multilingual information in a single data dump. DBpedia and Wikipedia provide language-specific dumps, so that we collect the dumps for the languages of interest, i.e. EN, FR, DE, RU and PT.  
The Wikipedia Current Events Portal is currently available in English only.
The mapping from the Wikidata identifiers to the Wikipedia and DBpedia identifiers 
required for the integration is collected as part of the Wikidata dump.
}

\ed{
As part of the pre-processing, the following information is created 
for each language\footnote{To obtain a complete list of the manually defined terms, expressions and mappings adopted in this work, please see the readme file in the open source software release provided at: \url{https://github.com/sgottsch/eventkg}}:
}

\begin{itemize}
    \item \textit{Terms:} Terms is a set of terms and regular expressions used throughout the extraction process. This includes the month names, weekday names, a black list of namespaces and prefixes of the Wikipedia articles to be ignored (e.g. the prefix ``Chronological\_list\_of\_'' in English) as well as regular expressions to detect titles of the Wikipedia articles representing events.
    \item \textit{Date expressions:} To extract dates from unstructured reference sources, a set of regular expressions is created. These expressions are sorted in the decreasing order of specificity, where time intervals are considered to be more specific than the individual dates or months. 
    For example, a specific regular expression to extract a span of two dates in English is: \texttt{@regex\-Month\-Day1@\allowbreak @hyphens\-Or@\allowbreak@regex\-Month\-Day2@}, where \texttt{@regex\-Month\-Day1@} denotes a month name followed by a date and \texttt{@hyphens\-Or@} is any kind of hyphen.
    This regular expression can match textual patterns such as ``February 17 --- April 23''. 
    A less specific expression is \texttt{@regex\-Day1@} that only checks for day numbers such as ``17''.
    Moreover, regular patterns to identify Wikipedia event lists such as ``2007 in Science'' are created, together with the rules to extract the temporal scope (the year 2007 in this example).
 \ed{
    \item \textit{Mapping of predicates representing event relations:} 
    We define a mapping table to identify predicates that 
    represent equivalent event relations in EventKG and its reference sources such as 
    \schema{so:has\-Sub\-Event} and Wikidata's ``\schema{part of}'' property.
    Examples of such mappings are shown in Table \ref{tab:mapping}.
    In this work we define the predicate mappings manually. 
    In future work 
    schema mapping techniques can be adopted to determine such links automatically. 
}
    
\end{itemize}

\textit{Identification and Extraction of Events}: Event instances are 
identified in the reference sources and extracted as follows:

\begin{itemize}[leftmargin=*]
\item 
\textbf{Wikidata} \cite{Erxleben:2014}: We identify events as subclasses of Wikidata's ``event'' (representing temporary and scheduled events like festivals or competitions) and ``occurrence'' (representing happenings like wars or ceremonies). 
Some of the  identified subclasses are blacklisted manually. For example, the class ``song'' is blacklisted because of the subclass hierarchy \schema{song} \textgreater\ \schema{musical form} \textgreater\ \schema{art form} \textgreater\ \schema{format}\ \textgreater\ \schema{arrangement} \textgreater\ \schema{act}  \textgreater\ \schema{process} \textgreater\ \schema{occurrence}. 
\item 
\textbf{DBpedia} \cite{dbpedia-swj}: For each language edition, we identify DBpedia events as instances of \schema{dbo:Event} or its subclasses.
\item 
\textbf{YAGO} \cite{Mahdisoltani:2014}: We do not use the YAGO ontology for event identification due to the noisy event subcategories we observed (e.g. \schema{event} \textgreater\ \schema{act} \textgreater\ \schema{activity} \textgreater\ \schema{protection}\ \textgreater\ \schema{self-defense} \textgreater\ \schema{martial\_art}). 
\item 
\textbf{Wikipedia}: We use Wikipedia category names that match a manually defined language-dependent regular expression (e.g. English category names that end with `` events'') as an indication that a knowledge graph entry linked to such an article is an event. 
\item 
\ed{\textbf{Wikipedia Event Lists}: For each language, we identify Wikipedia event lists by adopting a set of regular expressions defined manually during pre-processing. 
This way, Wikipedia pages with titles such as ``2007 in Science'' and ``August 11'' are retrieved. Within these pages, textual descriptions of events are collected using methods similar to \cite{Hienert:2012}. Using the ordered list of regular temporal expressions and Wikipedia link markup, representations of events including their descriptions, 
linked entities and dates are extracted.}
\item 
\textbf{WCEP}: In the Wikipedia Current Events Portal, events are represented through rather brief textual descriptions and refer to daily happenings. We extract WCEP events using the WikiTimes tool \cite{Tran:2014}.
\end{itemize}

\revA{
\begin{table*}[]
	\centering
    \footnotesize
    	\caption{Example property mapping between EventKG and its reference sources.}
	\label{tab:mapping}
	\begin{tabular}{|l||l|l|l|}
       \hline
\textbf{EventKG}  & \textbf{Wikidata} & \textbf{DBpedia}  & \textbf{YAGO} \\ \hline \hline
sem:hasPlace & \begin{tabular}[c]{@{}l@{}}wd:P276 (location)\\ wd:P30 (continent) \\ \ldots \end{tabular} & dbo:place & \begin{tabular}[c]{@{}l@{}} yago:isLocatedIn \\ yago:happenedIn\end{tabular} \\ \hline

sem:hasBeginTimeStamp & \begin{tabular}[c]{@{}l@{}} wd:P580 (start time) \\ wd:P585 (point in time) \\ wd:P1619 (date of official opening) \\ \ldots \end{tabular} & --- & \begin{tabular}[c]{@{}l@{}} yago:startedOnDate \\ yago:happenedOnDate \end{tabular}  \\ \hline

sem:hasEndTimeStamp & \begin{tabular}[c]{@{}l@{}} wd:P582 (end time) \\ wd:P585 (point in time) \\ \ldots \end{tabular} & --- & \begin{tabular}[c]{@{}l@{}} yago:endedOnDate \\ yago:happenedOnDate \end{tabular} \\ \hline

\schema{so:has\-Sub\-Event} & wd:P361 (part of) & \begin{tabular}[c]{@{}l@{}} dbo:isPartOf \\ dbo:isPartOfMilitaryConflict  \\ \ldots \end{tabular} & --- \\ \hline

\schema{so:previous\-Event} & wd:P155 (follows) & \begin{tabular}[c]{@{}l@{}} dbo:previousEvent \\ dbo:previousWork \end{tabular} & --- \\ \hline

\schema{so:next\-Event} & wd:P156 (followed by) & \begin{tabular}[c]{@{}l@{}} dbo:followingEvent \\ dbo:subsequentWork \end{tabular} & --- \\ \hline

\schema{so:con\-tained\-In\-Place} & \begin{tabular}[c]{@{}l@{}} wd:P36 (capital) \\ wd:P706 (located on terrain feature) \\ \ldots \end{tabular} & ---  & --- \\ \hline
	\end{tabular}
\end{table*}
}

\textit{Extraction of Event and Entity Relations}:
We extract the following types of relations:
1) \textit{Relations with temporal validity} are identified based on the availability of temporal information. 
Temporal relations are extracted from YAGO and Wikidata. DBpedia does not provide such information.
2) \textit{Relations with indirect temporal information}:
we extract all relations involving events as well as relations of entities with known existence time.
3) \textit{Other event and entity relations}: we use the manually defined mapping table shown in Table \ref{tab:mapping} to identify predicates that represent event relations in EventKG such as 
\schema{so:has\-Sub\-Event} (e.g. we map Wikidata's ``\schema{part of}'' property (P361) to \schema{so:has\-Sub\-Event} in cases where the property is used to connect events), \schema{dbo:previous\-Event} and \schema{dbo:next\-Event} as well as \schema{so:con\-tained\-In\-Place} to extract location hierarchies. 
\ed{4) \textit{Relation strength and event popularity information}: 
For each event-entity relation we extract language-specific interlinking information from Wikipedia. 
In particular we extract the number of links and the number of mentions 
for each relation involving events. 
Link and mentions are extracted from each Wikipedia language edition by parsing all of its pages.}

\textit{Integration}:
The statements extracted from the reference sources are included in the named graphs, such that each named graph corresponds to a reference source.
In addition, we create a named graph \schema{eventKG-g:event\_kg}
containing information resulting from integration and fusion.
Each \schema{sem:Event} and \schema{sem:Core} instance in the \schema{eventKG-g:event\_kg} graph integrates event-centric and entity-centric information from the reference sources related to equivalent real-world instances. 

\ed{
The integration of entities and events obtained from knowledge graphs and Wikipedia articles is conducted using existing \schema{owl:sameAs} links, as provided by the Wikidata dataset.
In particular, the entities and events covered by YAGO and different language versions of DBpedia and Wikipedia are also present in Wikidata. 
We use \schema{owl:sameAs} links to the Wikidata identifiers to represent each resource that is linked as equivalent in multiple reference sources as one resource in EventKG. That way, information regarding this resource in different reference sources, e.g. labels in different languages, is integrated.
In the current version of EventKG, we do not apply any entity resolution techniques to identify missing \schema{owl:sameAs} links in these reference sources. This can be addressed in future work to further increase the degree of integration. 

The events in the Wikipedia event lists and WCEP do not possess unique identifiers.
Such events are integrated using a rule-based approach to identify equivalent events.
Two events $e_1$ and $e_2$ extracted from such sources 
are represented as one EventKG event if the times of these events are identical ($e_1.time = e_2.time$) and the set of entities they link to overlaps. 
A special case is given if an event $e_1$ without an identifier links to an exactly one event $e_n$ with a known identifier and their times are equal. In that case, the text of $e_1$ is added as a description to $e_n$.
}

\textit{Fusion}: 
In the fusion step, we aggregate temporal, spatial and type information of \schema{eventKG-g:\allowbreak event\_kg} events using a rule-based approach. 

\begin{itemize}[leftmargin=*]

\item \textit{Time Fusion}: For each entity, event or relation with a known existence or a validity time stamp, time fusion is conducted using the following rules: (i) ignore the dates at the beginning or end of a time unit (e.g. January, 1st), if alternative dates are available; (ii) apply majority voting among the reference sources; (iii) take the time stamp from the more trusted source (in order: Wikidata, DBpedia, Wikipedia, WCEP, YAGO).

\item \textit{Location Fusion}: For each event in \schema{eventKG-g:\allowbreak event\_kg}, we take the union of its locations from the different reference sources and exploit the \schema{so:con\-tained\-In\-Place} relations to reduce this set to the minimum (e.g. the set \{Paris, France, Lyon\} is reduced to \{Paris, Lyon\}, while France can still be induced as a location using \schema{so:con\-tained\-In\-Place} transitively). 

\item \textit{Type Fusion}: We provide
 \schema{rdf:type} information according to the DBpedia ontology (dbo), using types and \schema{owl:sameAs} links in the reference sources.
 
\end{itemize}

\textit{Output}: Finally, extracted instances and relations are represented in RDF according to the EventKG data model (see Section \ref{sec:model}). 
As described above, information extracted from each reference source and the results of the fusion step are provided in separate named graphs.

%% file: 04d_running_example.tex
\subsection{Running Example: Barack Obama}
\label{ex:data}

In the context of our running example, we now provide an exemplary overview of the EventKG generation pipeline and illustrate how exemplar relations are expressed in the EventKG model and in the TKG. We refer to individual heterogeneous instances in the input data that are not yet expressed in the EventKG schema as \textit{data items}. Table \ref{tab:obama_example_data_collection} provides exemplary data items involving Barack Obama obtained from Wikidata, YAGO and different language editions of Wikipedia and DBpedia.

\begin{table*}[]
\centering
\caption{Example data items about Barack Obama extracted from different reference sources.}
\label{tab:obama_example_data_collection}
\begin{tabular}{|c||l|p{6.3cm}|p{6.1cm}|}
\hline
\textbf{\#} & \multicolumn{1}{|c|}{\begin{tabular}[c]{@{}l@{}}\textbf{Reference}\\\textbf{Source}\end{tabular}} & \multicolumn{1}{|c|}{\textbf{Data Item}} & \multicolumn{1}{|c|}{\textbf{Related Data Items}} \\ \hline \hline

1 & Wikipedia\textsubscript{EN} & \textit{8 May 2018}: President Trump announces his intention to withdraw the United States from the Iranian nuclear agreement. In a statement, former U.S. President Barack Obama calls the move "a serious mistake". & — \\ \hline

2 & Wikidata & \textit{Barack Obama}, \textit{significant event}, \textit{first inauguration of Barack Obama} & \begin{tabular}[c]{@{}p{6.1cm}@{}}Wikidata: \textit{first inauguration of Barack Obama}, \textit{point in time}, \textit{20 January 2009}\\    YAGO: \textit{first inauguration of Barack Obama}, \textit{was created on}, \textit{17 July 1981}\\    Wikidata: \textit{first inauguration of Barack Obama}, \textit{instance of}, \textit{United States presidential inauguration}\\    Wikidata: \textit{United States presidential inauguration}, \textit{subclass of*}, \textit{occurrence}\end{tabular}  \\ \hline

3 & Wikidata & \begin{tabular}[c]{@{}l@{}}\textit{Barack Obama}, \textit{spouse}, \textit{Michelle Obama}\\    \ \ \ \textit{start time}: \textit{3 October 1992}\end{tabular} & — \\ \hline

4 & DBpedia\textsubscript{FR} & \textit{Barack Obama}, \textit{prop-fr:candidat}, \textit{Élection présidentielle américaine de 2012} & \begin{tabular}[c]{@{}p{6.1cm}@{}}DBpedia\textsubscript{FR}: \textit{ Élection présidentielle américaine de 2012} \textit{owl:sameAs} \textit{United States presidential election, 2012}\\    Wikidata: \textit{United States presidential election, 2012}, \textit{point in time}, \textit{6 November 2012}\end{tabular} 
 \\ \hline

5 & Wikipedia\textsubscript{PT} & \textit{[The Portuguese Wikipedia page of Barack Obama links to the page ``Death of Osama bin Laden'' once.]} & Wikidata: \textit{Death of Osama bin Laden}, point in time, \textit{2 May 2011} \\ \hline

\end{tabular}
\end{table*}

\sg\textit{Identification and Extraction of Events.} The first data item is extracted from the English Wikipedia event list in the article ``2018 in the United States''. The entities ``first inauguration of Barack Obama'', ``United States presidential election, 2012'' and ``Death of Osama bin Laden'' from the data items \#2, \#3 and \#5 are identified as events using the class hierarchies in the reference sources. In this example, Obama's first inauguration is identified as an event, because it is an instance of ``United States presidential inauguration'', which can be tracked back to \schema{inauguration} \textgreater\ \schema{key event} \textgreater\ \schema{occurrence} in Wikidata.
Thus, the text event from data item \#1 and the event ``first inauguration of Barack Obama'' are stored as event instances with additional values such as a textual description for the former and a title for the latter event.

\sg{\textit{Extraction of Event and Entity Relations.} Given the set of events, we can now detect relations between them and other entities. For example, the statement that Barack Obama was involved in his own inauguration as US president is extracted from Wikidata. This statement represents an indirect temporal relation, as it alone does not provide the required temporal validity information, which needs to be extracted from a related fact about the event. Similarly, we can extract the information that Barack Obama was a candidate of the US elections in 2012 from the French DBpedia.}

With the help of Wikipedia links, we connect Barack Obama to the death of Osama bin Laden (data item \#5). Given the relation \texttt{?rel} that links to Barack Obama as the subject and to the event ``Death of Osama bin Laden'' as the object, the link information is modelled as follows, using a named graph (where \schema{eventKG-r:entity\_11973762} \sgg{represents Barack Obama and} \schema{eventKG-r:event\_527087} represents the event ``Death of Osama bin Laden''):

\begin{Verbatim}[samepage=true]
  ?rel rdfs:type
        eventKG-s:Relation .
  ?rel rdf:subject
        eventKG-r:entity_11973762 .
  ?rel rdf:object
        eventKG-r:event_527087 .

  eventKG-g:wikipedia_pt {
    ?rel eventKG-s:links 1 .
  } .
\end{Verbatim}

\ed{For the relation \texttt{?rel}, link information can be added using specific named graphs. For example, such information can model the co-mentions of Barack Obama and the death of Osama bin Laden in the Portuguese Wikipedia.}


Another type of information is coming from the temporal relations between two temporal entities: Here, the \textit{spouse} relation between Barack and Michelle Obama is directly assigned a temporal validity time by Wikidata.

\sg{\textit{Integration.} The entities ``Èlection présidentielle américaine de 2012'' and ``United States presidential election, 2012,'' are modeled as the same event resource in EventKG, using DBpedia's \schema{owl:sameAs} link.}

\sg{\textit{Fusion.} There are two different dates provided for the first inauguration of Barack Obama (data item \#2). While both dates are stored in EventKG together with their provenance information (i.e. as named graphs for Wikidata and YAGO), a single happening time for that event is created with our rule-based fusion approach (see Section \ref{sec:extraction}). As the majority voting is not sufficient here, we take the date from the higher trusted source. In this case, Wikidata's date (20 January 2009) is selected for EventKG's named graph.}

With that time information, the indirect temporal relation about Obama's participation in his own inauguration can be transformed into the following temporal relation in the $TKG$ generated from the named graph \schema{eventKG-g:event\_kg}:

\begin{Verbatim}[samepage=true]
Barack Obama,
significant event:
first inauguration of Barack Obama
[2009-01-20,2009-01-20]
\end{Verbatim}

%% file: 06a_characteristics.tex
\section{EventKG Characteristics \& Evaluation}
\label{sec:char-eval}

\sg{To demonstrate the quality of the data extraction, integration and fusion steps, we first show characteristics of EventKG and provide several comparisons to its reference sources in Section \ref{sec:characteristics}. Then, we provide evaluation results based on user annotations in Section \ref{sec:eval-eventkg}.}

\subsection{Characteristics}
\label{sec:characteristics}

In EventKG V1.1, we extracted event representations and relations in five languages -- English (EN), German (DE), French (FR), Russian (RU) and Portuguese (PT) -- from the latest available versions of each reference source as of 12/2017. EventKG uses open standards and is publicly available under a persistent URI\footnote{\url{https://doi.org/10.5281/zenodo.1112283}} under the CC BY 4.0 license\footnote{\url{https://creativecommons.org/licenses/by/4.0/}}.
Our extraction pipeline is available as open source software on GitHub\footnote{\url{https://github.com/sgottsch/eventkg}} under the MIT License\footnote{\url{https://opensource.org/licenses/MIT}}. A description of EventKG and example SPARQL queries are online\footnote{\url{http://eventkg.l3s.uni-hannover.de/}}. 
Two example SPARQL queries are also presented in Appendix \ref{sec:example-queries}.

Table \ref{tab:eventkg_overview} summarises selected statistics from the EventKG V1.1, released in 03/2018. Overall, this version provides information for over $690$ thousand events and over $2.3$ million temporal relations.  
Nearly half of the events ($46.75\%$) originate from the existing knowledge graphs; the other half ($53.25\%$) is extracted from semi-structured sources. 
The data quality of the individual named graphs directly corresponds to the quality of the reference sources. 
In \schema{eventKG-g:event\_kg}, the majority of the events ($76.21\%$) possess a known start or end time. 
Locations are provided for $12.21\%$ of the events. The coverage of locations can be further increased in future work, e.g. using NLP techniques to extract locations from event descriptions.
Along with over $2.3$ million temporal relations, EventKG V1.1 includes relations between events and entities for which the time is not available. This results in overall over 88 million relations. 
Approximately half of these relations possess interlinking information.

\begin{table*}[!t]
\centering
\footnotesize
\caption{Number of events and relations in \schema{eventKG-g:event\_kg}.}
\label{tab:eventkg_overview}
\begin{tabular}{l||r|r|r|}
\cline{2-4}
 & \multicolumn{1}{l|}{\textbf{\#Events}} & \multicolumn{1}{l|}{\textbf{Known time}} & \textbf{Known location} \\ \hline \hline
\multicolumn{1}{|l||}{Events from KGs} & 322,669 & 163,977 & \multicolumn{1}{r|}{84,304} \\ \hline
\multicolumn{1}{|l||}{Events from semi-structured sources} & 367,578 & 362,064 & not extracted \\ \hline
\multicolumn{1}{|l||}{Relations} & 88,473,111 & 2,331,370 & not extracted\\ \hline
\end{tabular}
\end{table*}

\subsubsection{Comparison of EventKG to its Reference Sources}
\label{sec:comparison}

We compare EventKG to its reference sources in terms of the number of identified events and completeness of their representations.
The results of the \sgg{event identification and extraction step} in Section \ref{sec:extraction} are shown in Table \ref{tab:event_identification}. 
EventKG with $690,247$ events contains a significantly higher number of events than any of its reference sources. 
This is especially due to the integration of knowledge graphs and semi-structured sources.

\begin{table*}[!t]
\scriptsize
\centering
\caption{Number of events identified in extracted from the reference sources.}
\label{tab:event_identification}
\begin{tabular}{l|c|c|c|c|c|c|c|c|c|c|l}
\cline{2-11}
 & \multicolumn{5}{c|}{\textbf{DBpedia}} & \multicolumn{5}{c|}{\textbf{Wikipedia event lists}} &  \\ \cline{1-1} \cline{12-12} 
\multicolumn{1}{|c|}{\textbf{Wikidata}} & \textbf{EN} & \textbf{FR} & \textbf{DE} & \textbf{RU} & \textbf{PT} & \textbf{EN} & \textbf{FR} & \textbf{DE} & \textbf{RU} & \textbf{PT} & \multicolumn{1}{c|}{\textbf{WCEP}} \\ \hline \hline
\multicolumn{1}{|r|}{266,198} & \multicolumn{1}{r|}{60,307} & \multicolumn{1}{r|}{43,495} & \multicolumn{1}{r|}{9,383} & \multicolumn{1}{r|}{5,730} & \multicolumn{1}{r|}{14,641} & \multicolumn{1}{r|}{131,774} & \multicolumn{1}{r|}{110,879} & \multicolumn{1}{r|}{21,191} & \multicolumn{1}{r|}{44,025} & \multicolumn{1}{r|}{18,792} & \multicolumn{1}{r|}{61,382} \\ \hline
\end{tabular}
\end{table*}

Table \ref{tab:named_events_comparison} presents a comparison of the event representations in EventKG and 
its reference knowledge graphs (Wikidata, YAGO, DBpedia).
As we can observe, through the integration of event-centric information, EventKG: 
1) enables better event identification (e.g. we can map $322,669$ events from EventKG to Wikidata, 
whereas only $266,198$ were identified as events in Wikidata initially - see Table \ref{tab:event_identification}) and 
2) provides more complete event representations (i.e. EventKG provides a higher percentage of events with specified temporal and spatial information compared to Wikidata, that is the most complete reference source). 
The most frequent event types are source-dependent (see Table \ref{tab:event_types}).

\begin{table*}[!t]
\centering
\footnotesize
\caption{Comparison of the event representation completeness in the source-specific named graphs (\sgg{after integration}).}
\label{tab:named_events_comparison}
\begin{tabular}{lrrr|r|r|r|r|r|}
\cline{5-9}
 & \multicolumn{1}{l}{} & \multicolumn{1}{l}{} & \multicolumn{1}{l|}{} & \multicolumn{5}{c|}{\textbf{DBpedia}} \\ \cline{2-4}
\multicolumn{1}{l|}{} & \multicolumn{1}{c||}{\textbf{EventKG}} & \multicolumn{1}{c|}{\textbf{Wikidata}} & \multicolumn{1}{c|}{\textbf{YAGO}} & \multicolumn{1}{c|}{\textbf{EN}} & \multicolumn{1}{c|}{\textbf{FR}} & \multicolumn{1}{c|}{\textbf{DE}} & \multicolumn{1}{c|}{\textbf{RU}} & \multicolumn{1}{c|}{\textbf{PT}} \\ \hline \hline
\multicolumn{1}{|l||}{\textbf{\#Events with}} & \multicolumn{1}{r|}{322,669} & \multicolumn{1}{r|}{322,669} & 222,325 & 214,556 & 78,527 & 62,971 & 47,304 & 35,682 \\ \hline
\multicolumn{1}{|l||}{\ Location (L)} & \multicolumn{1}{r|}{26.13\%} & \multicolumn{1}{r|}{11.70\%} & 26.61\% & 6.21\% & 8.32\% & 4.03\% & 10.60\% & 6.15\% \\ \hline
\multicolumn{1}{|l||}{\ Time (T)} & \multicolumn{1}{r|}{50.82\%} & \multicolumn{1}{r|}{33.00\%} & 39.02\% & 7.00\% & 17.21\% & 2.00\% & 1.35\% & 0.08\% \\ \hline
\multicolumn{1}{|l||}{\ L\&T} & \multicolumn{1}{r|}{21.97\%} & \multicolumn{1}{r|}{8.83\%} & 19.02 \% & 4.29\% & 0.00\% & 4.84\% & 1.18\% & 0.08\% \\ \hline
\end{tabular}
\end{table*}

\begin{table*}[!t]
 \centering
  \footnotesize
 \caption{The most frequent event types extracted from the references sources and the percentage of the events in that source with the respective type.
 }
 \label{tab:event_types}
 \begin{tabular}{lc|c|c|c|c|c|}
 \cline{3-7}
  & \multicolumn{1}{l|}{} & \multicolumn{5}{c|}{\textbf{DBpedia}} \\ \cline{2-2}
 \multicolumn{1}{l||}{} & \textbf{Wikidata} & \textbf{EN} & \textbf{FR} & \textbf{DE} & \textbf{RU} & \textbf{PT} \\ \hline \hline
 \multicolumn{1}{|l||}{\textbf{dbo:type}} & season & \begin{tabular}[c]{@{}c@{}}Military\\ Conflict\end{tabular} & \begin{tabular}[c]{@{}c@{}}Sports\\ Event\end{tabular} & \begin{tabular}[c]{@{}c@{}}Tennis\\ Tournament\end{tabular} & \begin{tabular}[c]{@{}c@{}}Military\\ Conflict\end{tabular} & \begin{tabular}[c]{@{}c@{}}Soccer\\ Tournament\end{tabular} \\ \hline
 \multicolumn{1}{|l||}{\textbf{Events, \%}} & \multicolumn{1}{r|}{11.37\%} & \multicolumn{1}{r|}{6.31\%} & \multicolumn{1}{r|}{21.86\%} & \multicolumn{1}{r|}{33.00\%} & \multicolumn{1}{r|}{11.87\%} & \multicolumn{1}{r|}{16.17\%} \\ \hline
 \end{tabular}
 \end{table*}

\subsubsection{Relation \& Fusion Statistics}
\label{sec:statistics}

Over 2.3 million temporal relations are an essential part of EventKG. 
The majority of the frequent predicates in EventKG such as ``member of sports team'' (882,398 relations), ``heritage designation'' (221,472), ``award received'' (128,125) and ``position held'' (105,333) originate from Wikidata. The biggest fraction of YAGO's temporal relations have the predicate ``plays for'' (492,263), referring to football players. Other YAGO predicates such as ``has won prize'' are less frequent. 
Overall, about $93.62\%$ of the temporal relations have a start time from 1900 to 2020.
$81.75\%$ of events extracted from knowledge graphs are covered by multiple sources.
At the fusion step, we observed that 93.79\% of the events that have a known start time agree on the start times across the different sources.

\subsubsection{Textual Descriptions}
\label{sec:lang_stat}

EventKG V1.1 contains information in five languages. 
Overall, $87.65\%$ of the events extracted from knowledge graphs provide an English label whereas only a small fraction ($4.49\%$) provide labels in all languages. 
Among the $367,578$ events extracted from the semi-structured sources, just $115$ provide a description in all five languages, e.g. the first launch of a Space Shuttle in 1981.
This indicates potential for further enrichment of  multilingual event descriptions in future work.

%% file: 06b_ekg_evaluation.tex
\subsection{Evaluation of EventKG}
\label{sec:eval-eventkg}

The aim of the evaluation is to assess the effectiveness of the event identification, 
time fusion and location fusion steps of the pipeline. 

\subsubsection{Event Identification}

We manually evaluated a random sample of the events identified in the event identification step of EventKG (Section \ref{sec:extraction}). For each reference source, we randomly sampled 100 events and manually annotated whether they represent real-world events or not. The results are shown in Table \ref{tab:user_evaluation_event_identification}.

For DBpedia and Wikidata, where we rely on the event types and type hierarchies, we achieve a precision of 98\% on average. On a random sample of 100 events extracted from the category names in the English and the Russian Wikipedia, we achieve 94\% and 88\% precision, correspondingly. One example for an entity wrongly identified as an event is the canceled project ``San Francisco Municipal Wireless'', which was part of the ``Cancelled projects and events'' category in Wikipedia.

\begin{table*}
\centering
\caption{User-evaluated precision for the identification of events with selected reference sources.}
\label{tab:user_evaluation_event_identification}
\begin{tabular}{l||l|l|l|l|l|l|}
\cline{2-7}
 & \textbf{Wikidata} & \textbf{DBpedia\textsubscript{DE}} & \textbf{DBpedia\textsubscript{RU}} & \textbf{DBpedia\textsubscript{PT}} & \textbf{Wikipedia\textsubscript{EN}} & \textbf{Wikipedia\textsubscript{RU}} \\ \hline \hline
\multicolumn{1}{|l||}{\textbf{Precision}} & \multicolumn{1}{r|}{96\%} & \multicolumn{1}{r|}{100\%} & \multicolumn{1}{r|}{100\%} & \multicolumn{1}{r|}{98\%} & \multicolumn{1}{r|}{94\%} & \multicolumn{1}{r|}{88\%} \\ \hline
\end{tabular}
\end{table*}

\subsubsection{Time Fusion}
\label{sec:time-fusion-results}

To evaluate the quality of the proposed rule-based time fusion approach, we randomly sampled $100$ events from EventKG, where each event has at least two reference sources that differ in the event happening time (i.e. start and/or end time). Three users have annotated this sample by providing a start and end time for at least 20 events each. Additionally, we asked the users to denote which source they used to research the actual event dates. For our evaluation, we then checked how many of the user-given start and end dates are available in the reference sources and the joint EventKG named graph, and we computed how many of these dates are correct with respect to the user annotations.

\begin{table*}[]
\centering
\footnotesize
\caption{Evaluation of EventKG's time information. For EventKG and the reference sources, the percentage of correct, wrong and missing event dates with respect to the user annotations in our sample is shown. These are based on the random sample of events where the reference sources show disagreement between time information provided.}
\label{tab:ekg_evaluation_time}
\begin{tabular}{l||r|r|r|r|r|r|r|r|r|r|}
\cline{2-11}
 & \multicolumn{3}{c|}{\textbf{Start Dates}} & \multicolumn{3}{c|}{\textbf{End Dates}} & \multicolumn{4}{c|}{\textbf{Start and End Dates}} \\ \hline \hline
\multicolumn{1}{|l||}{\textbf{Source}} & \multicolumn{1}{l|}{\textbf{Correct}} & \multicolumn{1}{l|}{\textbf{Wrong}} & \multicolumn{1}{l|}{\textbf{Missing}} & \multicolumn{1}{l|}{\textbf{Correct}} & \multicolumn{1}{l|}{\textbf{Wrong}} & \multicolumn{1}{l|}{\textbf{Missing}} & \multicolumn{1}{l|}{\textbf{Correct}} & \multicolumn{1}{l|}{\textbf{Wrong}} & \multicolumn{1}{l|}{\textbf{Missing}} & \multicolumn{1}{l|}{\textbf{Precision}} \\ \hline
\multicolumn{1}{|l||}{EventKG} & \textbf{71} & 25 & 0 & \textbf{73} & 23 & 0 & \textbf{144} & 48 & 0 & \textbf{0.75} \\ \hline
\multicolumn{1}{|l||}{Wikidata} & 40 & 33 & 23 & 33 & 29 & 34 & 73 & 62 & 57 & 0.54 \\ \hline
\multicolumn{1}{|l||}{YAGO} & 21 & 60 & 15 & 20 & 57 & 19 & 41 & 117 & 34 & 0.26 \\ \hline
\multicolumn{1}{|l||}{DBpedia\textsubscript{EN}} & 12 & 5 & 79 & 13 & 4 & 79 & 25 & 9 & 158 & 0.74 \\ \hline
\multicolumn{1}{|l||}{DBpedia\textsubscript{DE}} & 0 & 2 & 94 & 2 & 0 & 94 & 2 & 2 & 188 & 0.5\\ \hline
\multicolumn{1}{|l||}{DBpedia\textsubscript{FR}} & 6 & 17 & 73 & 15 & 8 & 73 & 21 & 25 & 146 & 0.46 \\ \hline
\multicolumn{1}{|l||}{DBpedia\textsubscript{RU}} & 0 & 2 & 94 & 0 & 2 & 94 & 0 & 4 & 188 & 0 \\ \hline
\end{tabular}
\end{table*}

Table \ref{tab:ekg_evaluation_time} provides the result overview: As the time fusion does always adopt accessible time information from any reference source, all events in our random sample possess time information.
Wikidata and YAGO provide the next highest coverage of time information. 
In terms of precision, EventKG outperforms these two reference sources by $21\%$ (Wikidata) 
and $49\%$ (YAGO). This result confirms the quality of the proposed rule-based time fusion approach.
\ed{
The results of a McNemar's test \cite{McNemar:1947} has shown a two-tailed p-value of less than $0.0001$, which confirms the statistical significance of this result.
}

Table \ref{tab:ekg_evaluation_time_sources} provides an overview of the sources most often used for finding the event dates by the users participating in the evaluation. In $69\%$ of the cases, the users adopted Wikipedia articles in different languages as their source. When the users did not use Wikipedia, either the information presented on the search engine's result page ($18.5\%$ of the cases) or domain-specific web sites such as www.singapore-elections.com or www.un.org were used.

\begin{table}[]
\centering
\caption{Time Fusion Evaluation: The most frequent sources used by the users to lookup event start and end dates.}
\label{tab:ekg_evaluation_time_sources}
\begin{tabular}{|l||r|r|}
\hline
\textbf{Source}  & \textbf{\#Uses} & \textbf{Percentage}  \\ \hline \hline
en.wikipedia.org & 117 & 58.5\% \\ \hline
www.google.com & 37 & 18.5\% \\ \hline
de.wikipedia.org & 14 & 7.0\% \\ \hline
\textit{no source used} & 7 & 3.5\% \\ \hline
fr.wikipedia.org & 6  & 3.0\%\\ \hline
www.singapore-elections.com & 2 & 1.0\% \\ \hline
www.un.org & 2 & 1.0\% \\ \hline
\multicolumn{3}{|c|}{…} \\ \hline
\end{tabular}
\end{table}

\subsubsection{Location Fusion}

\sg{To evaluate the correctness of the extracted locations, we selected a random sample of $100$ events with at least one location. 
In case of locations, multiple correct values are possible, for example South America, the United States of Colombia and the Colombia-Ecuador border are valid locations for the Ecuadorian-Colombian War.
We presented all locations from each reference source to the users and for each location asked the users to verify whether that location is correct or not. Four users have annotated that sample. }

Table \ref{tab:ekg_evaluation_location} provides the result for our evaluation of the location fusion. We distinguish between the locations directly provided by EventKG and those which could be inferred using sub-location information via \schema{so:containedInPlace}. We refer to this extended knowledge graph as EventKG* throughout this evaluation. EventKG and EventKG* have by far the highest coverage of locations (EventKG* finds $78.13\%$ more event locations than YAGO and $159.10\%$ more than in Wikidata), while keeping the number of wrong locations low (approx. $7\%$), although it also inherits wrong locations as provided by the reference sources due to the adopted location fusion mechanism.
\ed{
The results of a McNemar's test \cite{McNemar:1947} has shown a two-tailed p-value of $0.0005$, which confirms statistical the significance of this result.
}

Table \ref{tab:ekg_evaluation_location_sources} lists the sources used by the users in this task. Similarly to the evaluation of the time fusion, Wikipedia and Google were the most frequently used sources, followed by domain-dependent ones such as kicker.de for locating football matches. However, in $26.51\%$ of the cases in this task, the users did not use a source at all, mainly because many event locations are self-explanatory or contained in the event names. For example, no source was needed to verify the locations Monaco and Circuit de Monaco for the 1956 Monaco Grand Prix.

\begin{table}[]
\centering
\caption{Evaluation of EventKG's location information. For each event in the sample, users judged for each location in EventKG and the reference sources whether it is correct.}
\label{tab:ekg_evaluation_location}
\begin{tabular}{|l||r|r|r|}
\hline
\textbf{Source} & \multicolumn{1}{l|}{\textbf{Correct}} & \multicolumn{1}{l|}{\textbf{Wrong}}& \multicolumn{1}{l|}{\textbf{Precision}} \\ \hline \hline
EventKG* & \textbf{116} & 7 & 94.31\% \\ \hline
EventKG & 87 & 4 & 95.60\% \\ \hline
YAGO & 64 & 2 & 96.97\% \\ \hline
Wikidata & 44 & 2 & 95.65\% \\ \hline
DBpedia\textsubscript{EN} & 15 & 1 & 93.75\% \\ \hline
DBpedia\textsubscript{FR} & 7 & 0 & 100.0\% \\ \hline
DBpedia\textsubscript{DE} & 1 & 0 & 100.0\% \\ \hline
DBpedia\textsubscript{RU} & 4 & 1 & 80.0\% \\ \hline
DBpedia\textsubscript{PT} & 3 & 1 & 75.0\%\\ \hline
\end{tabular}
\end{table}

\begin{table}[]
\centering
\caption{Location Fusion Evaluation: The most frequent sources used by the users to lookup event locations.}
\label{tab:ekg_evaluation_location_sources}
\begin{tabular}{|l||r|r|}
\hline
\textbf{Source}  & \textbf{\#Uses} & \textbf{Percentage} \\ \hline \hline
en.wikipedia.org   & 58 & 43.94\% \\ \hline
\textit{no source used} & 35 & 26.51\% \\ \hline
de.wikipedia.org & 7 & 5.3\% \\ \hline
www.google.com & 5 & 3.79\% \\ \hline
everipedia.org & 3 & 2.0 \% \\ \hline
fr.wikipedia.org & 3 &  2.0 \%                    \\ \hline
www.kicker.de & 2 & 1.51\%                     \\ \hline
\multicolumn{3}{|c|}{…} \\ \hline
\end{tabular}
\end{table}

%% file: 06c_ekg_v2.tex
\subsection{EventKG V2.0}
\label{sec:eventkgv2}

The characteristics, statistics and evaluation results presented 
in this article refer to EventKG V1.1 released in March 2018. 

In February 2019, we released EventKG V2.0 that includes  
a number of updates with respect to the: 
i) inclusion of the current content of the reference sources
and extended language coverage, 
ii) enhanced relation fusion, 
iii) inclusion of geographic information, and
iv) inclusion of information regarding temporal granularity. 
In the following we describe these extensions in more detail. 

\textbf{Reference sources and language coverage}. Event\-KG V2.0 includes data extracted from the reference sources presented in Section \ref{sec:extraction} as of January 1st, 2019. Furthermore, EventKG V2.0 includes Italian as the sixth language, in addition to the five languages supported in Event\-KG V1.1. Overall, this leads to $979,623$ events included in the dataset. 
    
\textbf{Relation fusion}. In Event\-KG V2.0 we performed fusion of \schema{eventKG-s:\allowbreak Relation} instances extracted from different reference sources based on property mappings and similarity. 
\schema{eventKG-s:\allowbreak Relation} instances are fused if the following conditions are met: 
(1) The values of \schema{rdf:subject}, \schema{rdf:object}, \schema{sem:\allowbreak hasBegin\-TimeStamp} and \schema{sem:\allowbreak hasEnd\-TimeStamp} are the same, and
(2) the \schema{sem:\allowbreak role\-Type} values are linked via existing \schema{owl:sameAs} relations in the reference sources. 
For example, this concerns properties such as ``place of birth'' (Wikidata), ``wasBornIn'' (YAGO) and ``birthPlace'' (English DBpedia).
    
\textbf{Geographic information}. For \schema{sem:\allowbreak Place} and \schema{sem:\allowbreak Event} instances, geographic coordinates available in the reference sources are added to EventKG V2.0. The coordinates are represented through their latitude and longitude as values of \schema{so:latitude} and \schema{so:longitude}.
    
\textbf{Temporal granularity information}.  
In Event\-KG V2.0 we enriched the dates encoded by 
\schema{sem:has\-Begin\-TimeStamp} and \schema{sem:has\-End\-TimeStamp}
with granularity information, which denotes the precision of a given date. 
To this end, the properties \schema{eventKG-s:start\-Unit\-Type} and \schema{eventKG-s:end\-Unit\-Type} are added to the schema. 
Their range is \schema{time:Temporal\-Unit}, which comprises existing classes 
in the Time Ontology\footnote{\url{http://www.w3.org/2006/time\#} (namespace prefix ``\schema{time:}'')} (\schema{time:unit\-Day}, \schema{time:\allowbreak unit\-Month} and \schema{time:\allowbreak unit\-Year}), as well as newly created classes (\schema{event\-KG-s:\allowbreak unit\-Decade} and \schema{event\-KG-s:\allowbreak unit\-Century}).
For example, the granularity information helps to identify whether the start time ``January 1st, 1981'' refers to that actual day (\schema{eventKG-s:\allowbreak start\-Unit\-Type} \schema{time:unit\-Day}) or to an unknown day of the year (\schema{eventKG-s:\allowbreak start\-Unit\-Type} \schema{time:\allowbreak unit\-Year}).

EventKG V2.0, its updated schema information and statistics are accessible online.\footnote{\url{http://eventkg.l3s.uni-hannover.de/}}

%% file: 05a_method.tex
\section{Generation of Biographical Timelines}
\label{sec:timelines}

In this section, we show how EventKG can be applied as a temporal knowledge graph for the task of biographical timelines generation.

\sg{First, we present our approach based on distant supervision in Section \ref{sec:approach}. The features used in the relevance model are introduced in Section \ref{sec:relevance_model}.
Subsequently, we describe the benchmarks involved in our process to generate biographical timelines in Section \ref{sec:benchmarks} and discuss how the model is used to generate them in Section \ref{sec:timeline_generation}. Finally, we illustrate these steps on our running example of Barack Obama's timeline in Section \ref{ex:timeline}.}

\subsection{Approach}
\label{sec:approach}

Given a timeline entity $e$ for which we need to generate a biographical timeline, the number of \textit{candidate timeline entries} (i.e. temporal relations involving $e$) is potentially very high, especially for popular entities and a large-scale temporal knowledge graph. In fact, for our set of popular persons described later in Section \ref{sec:benchmark}, EventKG contains 272.75 temporal relations per person entity on average. 
In order to determine the relevance of a temporal relation to the timeline entity we propose a classification approach using distant supervision. 
\begin{figure*}[t]
\centering
  \includegraphics[width=0.75\linewidth]{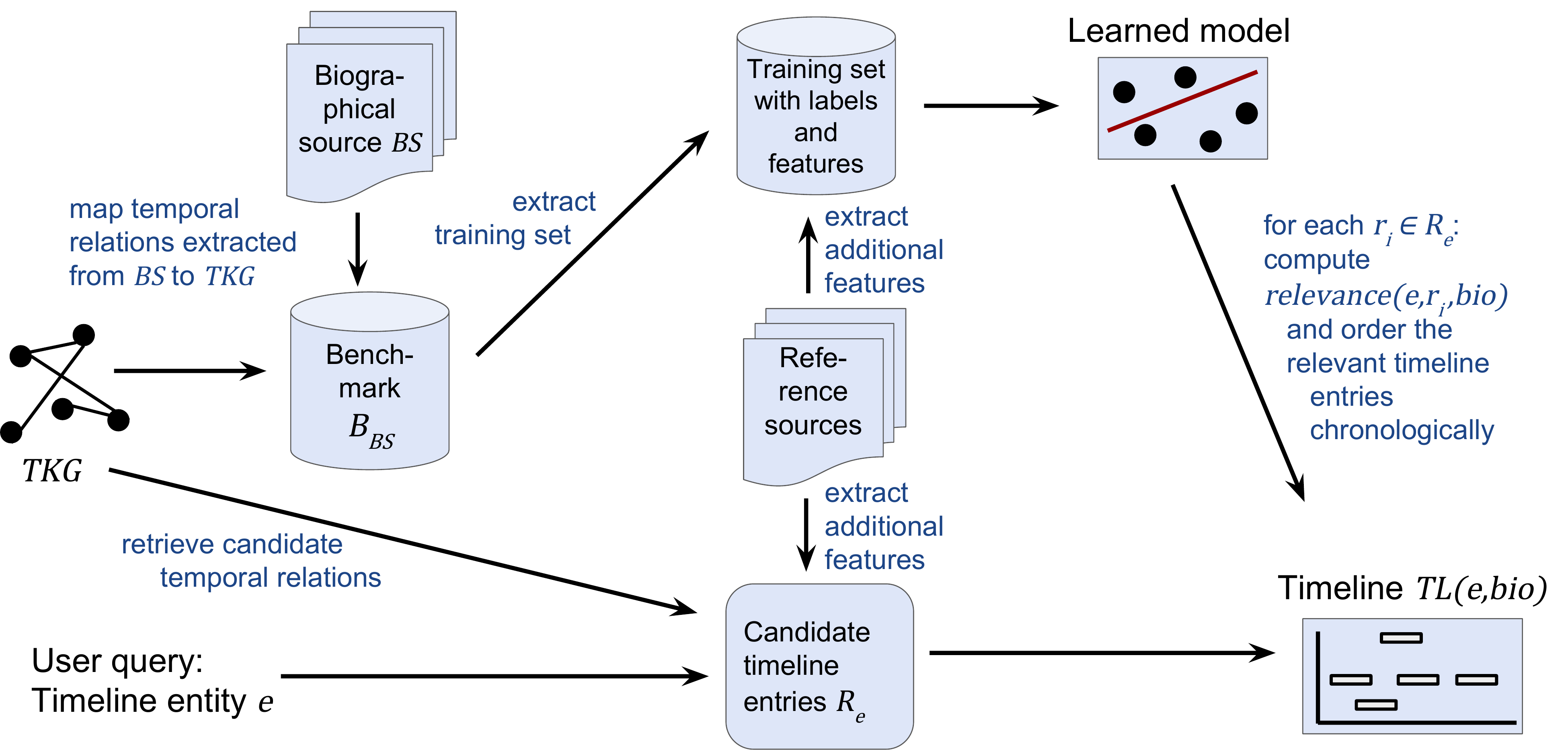}
  \caption{Creating a timeline for a timeline entity $e$, after training a model from a biographical source to predict the relevance of temporal relations in the TKG for biographical timelines.}
  \label{fig:diagram}
\end{figure*}
The key idea of our approach is to learn a relevance model for temporal 
relations using occurrences of these relations extracted from \textit{biographical sources}.  
Examples of such biographical sources include collections of biographical or encyclopedic articles. 
\ed{We adopt a distant supervision approach, where we assume that a particular temporal relation $r$ is relevant for the entity's biography if this relation occurs in a known biographical source. An overview of the training phase and the timeline generation is depicted in Figure \ref{fig:diagram}, which illustrates the role of the TKG, the biographical and reference sources and the benchmark.
Initially, we use the temporal knowledge graph and a biographical source to create a benchmark that provides relevance judgements for candidate timeline entries. We train the prediction model with features extracted for each candidate timeline entry. This includes entity type and interlinking information included in the named graphs corresponding to the reference sources of EventKG. 
To generate a timeline for a timeline entity $e$, we collect its candidate timeline entries $R_e$ from $TKG$ and identify the relevant entries using the trained model.}

\subsection{Relevance Model}
\label{sec:relevance_model}

\ed{
In our approach we train a classification model that identifies the relevance of a candidate timeline entry towards a biography of the timeline entity $e$. 
The candidate timeline entry is a temporal relation involving $e$ and obtained from a knowledge graph. 
To train such classification models, we adopt  
a range of features in several categories reflecting the characteristics of the timeline entity, the entity connected to it via a temporal relation, the temporal relation and time information. In total, we consider $4$ language-independent numerical features, $6$ language-dependent features, as well as a number of binary features representing frequent entity types and properties in EventKG.}

We illustrate the features described in the following
at the example of the candidate timeline entry representing Barack Obama's participation in his second inauguration (see Figure \ref{fig:relation_instances}) in Table \ref{tab:feature_example}.

\subsubsection*{Timeline Entity Features}
\label{sec:tef-features}

The timeline entity features (TEF) reflect specific characteristics of the timeline entity $e$.
These features address the intuition that the relevance of the particular temporal relation $r$ for a given timeline entity $e$ depends on the specific characteristics of $e$.
For example, winning an award may be more important for athletes or actors than for politicians. Based on this intuition, we introduce the timeline entity features:

\begin{itemize}
	\item[TEF-C] Timeline entity characteristics: A set of binary features denoting if the entity is an instance of the specific type (e.g. a politician or an actor). 
\end{itemize}

\subsubsection*{Connected Entity Features}
\label{sec:cef-features}

The connected entity features (CEF) take into account characteristics of the connected entity $e'$. 
In particular, we consider indications of the importance and popularity of $e'$ in the context of the reference collections by using 
mention counts, similar to Thalhammer et al. \cite{Thalhammer:2016}.
In particular, we consider different representations of the mention counts of $e'$. 

\begin{itemize}
	\item [CEF-M] Connected entity mentions: The set of features, each reflecting the absolute 
	number of mentions of the connected entity $e'$ in a reference collection.   
	\item [CEF-MR] Connected entity mentions rank: For each reference collection, we rank the entities connected to the timeline entity $e$ by the number of their mentions. This feature represents the rank of the specific connected entity, where the rank of $1$ is assigned to the entity with the highest number of mentions.  
	\item [CEF-MRR] Connected entity mentions relative rank: We normalise the CEF-MR rank by the maximal rank. 
	\item [CEF-E] Connected entity represents a real-world event: A binary feature denoting whether the connected entity is an event (i.e. $e'\in \mathcal{V}$).
\end{itemize}

\subsubsection*{Features of Temporal Relations}
\label{sec:trf-features}

The features of temporal relations (TRF) reflect semantics of the temporal relation between the timeline entity and the connected entity.
Furthermore, we consider features related to the importance and popularity of entity relations. 

\begin{itemize}
	\item [TRF-PI] Property identifier: Temporal relations possess property identifiers $r_{uri}$ that express semantics of the relation (e.g. \textit{dbo:spouse}). Each property identifier is modelled as a binary feature. 
	\item [TRF-M] Relation mentions: The number of co-mentions of both entities involved in the temporal relation in a reference collection (independent of relation semantics).
	\item [TRF-MR] Relation mentions rank: We rank the connected entities according to the number of their co-mentions with the timeline entity in a reference collection. This feature represents the rank of the specific connected entity involved in the relation.
    \item [TRF-MRR] Relation mentions relative rank: We normalise the TRF-MR rank by its maximal rank.
\end{itemize}

\subsubsection*{Temporal Features}
\label{sec:tf-features}

The temporal features (TF) reflect the relevance of the temporal relations based on the time information. 
This includes the temporal differences in the existence time of the entities or happening times of the events involved in the relation. 
For example, Barack Obama gave a speech related to World War II - a historical event finished before Obama's birth date in 1961. Here, the temporal difference in the existence times of both entities can be an indication of the low relevance of this speech for Obama's biography. 
Therefore, we attempt to learn to discard the temporal relations involving events that happened too early for the entity timeline. This had been also observed by Althoff et al. \cite{Althoff:2015} who implemented a rule to discard such relations. 
In addition to that, our temporal features could help to learn whether some events may be more relevant at specific stages of the entity's life or existence. Furthermore, temporal features include the provenance of the temporal information by denoting whether a relation was induced from an indirect temporal relation or not.

To capture this intuition, we introduce the following temporal features:
\begin{itemize}
	\item [TF-TDS] Temporal distance (start): The temporal distance between the beginning of the existence time of the timeline entity and the start of the relation validity time $e_{start}-r_{start}$. 
	\item [TF-TDE] Temporal distance (end): The same feature as TF-TDS, but using the entity existence end time $e_{end}-r_{start}$.
	\item [TF-TP] Time provenance: This categorical feature specifies the provenance of the relation validity time. If the relation has initially been a temporal relation, the feature value is set to $3$. If the temporal validity was induced from an event happening time ($e_j \in \mathcal{V}$), then the feature value is set to $2$; $1$ otherwise ($e_j \in \mathcal{E'}$).
\end{itemize}

\begin{table*}[t!]
\footnotesize
\centering
\caption{\sg{Selected feature values for the candidate timeline entry ``Barack Obama, significant event, Second inauguration of Barack Obama'' for the timeline entity ``Barack Obama''.}    }
\label{tab:feature_example}
\begin{tabular}{|l|l|r|p{10cm}|}
\hline
\multicolumn{1}{|c|}{\textbf{Feature}} & \multicolumn{1}{c|}{\textbf{\begin{tabular}[c]{@{}c@{}}Feature\\ Instance\end{tabular}}} & \multicolumn{1}{c|}{\textbf{Value}} & \multicolumn{1}{c|}{\textbf{Note}} \\ \hline
\multirow{3}{*}{\textbf{TEF-C}} & Politician & 1 & Barack Obama is an instance of dbo:Politician. \\ \cline{2-4} 
 & President & 1 & Barack Obama is an instance of dbo:President. \\ \cline{2-4} 
 & Scientist & 0 & Barack Obama is not an instance of dbo:Scientist. \\ \hline
\textbf{CEF-M} & CEF-M\textsubscript{EN} & $84$ & The inauguration is linked 84 times in the English Wikipedia. \\ \hline
\textbf{CEF-MR} & CEF-MR\textsubscript{EN} & $361$ & Among all entities connected to Obama in the English Wikipedia, the inauguration is linked the 361st most times. \\ \hline
\textbf{CEF-MRR} & CEF-MR\textsubscript{EN} & $0.817$ & Among all entities connected to Obama in the English Wikipedia, there are $442$ different CEF-MR\textsubscript{EN} scores, such that inauguration's relative rank is $\frac{361}{442}\approx 0.817$. \\ \hline
\textbf{CEF-E} & CEF-E & 1 & The inauguration is an instance of sem:Event. \\ \hline
\multirow{2}{*}{\textbf{TRF-PI}} & wd:significantEvent & 1 & Obama is connected to the inauguration through Wikidata's ``significant event'' property. \\ \cline{2-4} 
 & wd:spouse & 0 & Barack Obama is not connected to the inauguration through Wikidata's ``spouse'' property. \\ \hline
\textbf{TRF-M} & TRF-M\textsubscript{PT} & 4 & In the Portuguese Wikipedia, there are 4 sentences mentioning both Barack Obama and the inauguration. \\ \hline
\textbf{TRF-MR} & TRF-MR\textsubscript{PT} & 18 & Among all co-mentions of Barack Obama and an event, the co-mention with the inauguration is the 18th most frequent one the Portuguese Wikipedia. \\ \hline

\textbf{TRF-M} & TRF-M\textsubscript{ALL} & 36 & In all the five involved Wikipedia language editions together, there are 36 sentences mentioning both Obama and the inauguration. \\ \hline
\textbf{TRF-MR} & TRF-MR\textsubscript{ALL} & 39 & Among all co-mentions of Barack Obama and an event, the co-mention with the inauguration is the 39th most frequent one in all the five involved Wikipedias together. \\ \hline

\textbf{TF-TDS} & TF-TDS & 18798 & The inauguration started $18798$ days ($51$ years) after Barack Obama's birth. \\ \hline
\textbf{TF-TDE} & TF-TDE & 18798 & The inauguration ended $18798$ days ($51$ years) after Barack Obama's birth. \\ \hline
\textbf{TF-TP} & TF-TP & 2 & The validity time assigned to this temporal relation is induced from the happening time of an event instance. \\ \hline
\end{tabular}
\end{table*}

\subsection{Benchmarks for Distant Supervision}
\label{sec:benchmarks}
To facilitate supervised model training, we require a benchmark that provides relevance judgements for temporal relations. These judgements can be obtained from the specific biographical source.

\begin{definition}
A \emph{benchmark} $B$ is a mapping of the form: $relevance(e_i, r_j, bio) \mapsto J, J\in\{0,1\}$, 
where $e_i$ is a temporal entity, $r_j$ is a temporal relation involving $e_{i}$ and $J$ is a relevance judgement.
\end{definition}

Given the large number of entities and temporal relations in the existing knowledge graphs, manual relevance judgements appear unfeasible.
Therefore, we adopt an automatic approach to benchmark generation. 
We extract entities and temporal relations contained in the biographical sources and 
map them to the temporal relations in $TKG$ using an automatic procedure involving source-specific heuristics (described later in Section \ref{sec:benchmark}). 
Temporal relations extracted from the biographical sources are considered 
relevant.

Although the resulting benchmarks can potentially contain noisy relevance judgements due to the automatic extraction and mapping methods applied, our experimental results demonstrate that these benchmarks, used as a training set in a distant supervision method, facilitate generation of high quality timelines. 

The benchmarks created in this work are publicly available online\footnote{\url{http://eventkg.l3s.uni-hannover.de/timelines.html}}.

\subsection{Model Training and Timeline Generation}
\label{sec:timeline_generation}

We address the relevance estimation for a timeline relation $r$ with respect to the timeline entity $e$ as a classification problem. 
For each biographical source $BS$, we build a classification model using the features presented in Section \ref{sec:relevance_model} and a binary classifier. 

Note that a classification model is chosen over a ranking-based approach because of two reasons: 
First, the timeline entries are ordered chronologically and not by their importance. 
Therefore, for the purpose of timeline generation we can assume that each timeline entry is equally relevant. 
Second, if a ranked list of timeline entries would be provided, a cut-off threshold value would still be required to decide which of the entries 
are to be shown. 

To facilitate efficient training we limit the number of instances of the TEF-C and TRF-PI features considered. 
In particular, the $50\%$ most frequent types in the training set are added as a TEF-C feature.
Furthermore, only properties that occur in at least $25\%$ of the relations in the training set are added as a TRF-PI feature.

Our benchmark is equally divided into a training and a test set of person entities, so that the relevance judgements are obtained from the training set. We adopt a binary notion of relevance. 
The datasets used as biographical sources to build the classification models are presented in Section \ref{sec:benchmark}.

We use the resulting classification model to build a timeline $TL(e, bio)$.
Each candidate timeline entry (i.e. a temporal relation involving the timeline entity $e$  in $TKG$) is classified using the classification models learned from a biographical source. 
The classification function $relevance(e, r, bio)$ uses this model to classify the temporal relations of the timeline entity $e$ as  either $0$ (non-relevant) or $1$ (relevant). 
As illustrated in Figure \ref{fig:diagram}, the timeline is generated by ordering the timeline entries classified as relevant by their start time.

%% file: 05b_running_example.tex
\subsection{Running Example: Barack Obama}
\label{ex:timeline}

As discussed in Section \ref{ex:data}, EventKG contains many relations involving Barack Obama. In order to create a timeline of his life, we collect all relations with Obama as a subject or an object, together with their temporal validity. One example is the temporal relation about Obama's first inauguration shown at the end of Section \ref{ex:data}.

Due to the more than $2,500$ candidate timeline entries for Obama, we now need to apply the previously trained model to determine the timeline entries relevant for a biography. To this end, we train the classifier that predicts whether a candidate timeline entry is relevant given a biographical source, i.e. whether it is probable to be part of entity biography in such source. All candidate timeline entries that are classified as relevant by this model are inserted into the timeline in chronological order.

Figure \ref{fig:timeline_example} provides a visual representation of Obama's timeline obtained using a model trained on a Wikipedia abstracts dataset (BS-ENC) described later in Section \ref{sec:eval-timeline}.

%% file: 07a_dataset.tex
\section{Setup and Evaluation of the Biographical Timeline Generation}
\label{sec:eval-timeline}

\begin{table*}[ht]
	\centering
	\scriptsize
	\caption{Example data extracted from the biographical sources for Barack Obama.}
	\label{tab:benchmark_obama}
	\begin{tabular}{l||l|l|}
		\cline{2-3}
          & \multicolumn{1}{|c|}{\textbf{BS-BIO}} & \multicolumn{1}{|c|}{\textbf{BS-ENC}} \\ \hline \hline
		\multicolumn{1}{|l||}{\textbf{Source}} & \begin{tabular}[c]{@{}l@{}} biography.com,\\thefamouspeople.com\end{tabular} & Wikipedia\textsubscript{EN} abstracts \\ \hline
		\multicolumn{1}{|l||}{\textbf{\begin{tabular}[c]{@{}l@{}}Example\\Data\end{tabular}}} &
		\begin{tabular}[c]{@{}l@{}}1961-8-4, \{Honolulu\}\\1979, \{Punahou School, Basketball\}\\2000, \{Democratic Party, Bobby Rush\}\\2010-8, \{War in Afghanistan, Iraq\}\end{tabular} 
		 & \begin{tabular}[c]{@{}l@{}}1961, \{Honolulu\}\\2013, \{US presidential election 2012, Mitt Romney,\\\ \ \ Second inauguration of Barack Obama\}\\2009, \{Nobel Peace Prize\}\end{tabular} \\ \hline
	\end{tabular}
\end{table*}

In this section we first describe the biographical sources and the set of timeline entities used to create our biographical timeline benchmark used to train the classification models (Section \ref{sec:benchmark}) and to run our experiments described in Section \ref{sec:svm_setup}. Then, we evaluate our approach against a baseline (Sections \ref{sec:baseline} and \ref{sec:evaluation}). 

\subsection{Benchmark: Entities and Biographical Sources}
\label{sec:benchmark}

We collect a dataset $\mathcal{P}$ that contains $2,760$ timeline entities of the type \textit{Person}, including its subtypes like politicians, actors, musicians and athletes. 
This set of $2,760$ entities contains all persons that are included in EventKG and described in each biographical source described below.
\sgg{Consequently, the training and the test set consist of $1,380$ person entities each, after random division.}

\begin{samepage}
To train the relevance models for the biographical timeline generation, we consider the following biographical sources: 
 \begin{itemize}
\item BS-BIO: Biographical articles;
\item BS-ENC: Encyclopedic articles.
\end{itemize}
\end{samepage}

\subsubsection*{Biographical articles (BS-BIO):}

Biographies of important entities (e.g. famous people) are available in form of textual descriptions from dedicated Web sources.
We collect data from two publicly accessible biographical web sources (Thefamouspeople.com\footnote{\url{www.thefamouspeople.com}} and 
Biography.com\footnote{\url{www.biography.com}}). 
After collecting the biographical texts from both websites, they are preprocessed as follows: 
1) The texts are split into sentences using the Stanford Tokenizer \cite{Manning:2014}. 
2) Time expressions are collected from each sentence using HeidelTime \cite{Strotgen:2010}. 
3) Entity mentions are identified using DBpedia Spotlight \cite{Mendes:2011}. 
Table \ref{tab:benchmark_obama} illustrates example annotations in the \textit{BS-BIO} and \textit{BS-ENC} datasets extracted for the entity Barack Obama, including his birth, education and political activities.
In order to map the extracted information to the temporal relations in the $TKG$, we use the following rule-based approach: 
An annotated sentence in the biographical article is mapped to the temporal relation in $TKG$ if they both happened on exactly the same date, or if they share both entities and time. A special case is given if one of the linked entities is an event in $\mathcal{V}$. In that case, temporal overlap is not required, as events are typically inherently connected to a validity time span.
The mapped temporal relations from the $TKG$ are added to the $B_{BS-BIO}$ benchmark.

\subsubsection*{Encyclopedic articles (BS-ENC):}

Wikipedia is a rich source of encyclopedic information. 
Wikipedia articles usually provide an abstract - a brief overview of the specific entity (e.g. person's life) that typically contains important biographical sentences \cite{Chisholm:2017, Lebret:2016}. From these abstracts, we extract all the event mentions, i.e. links to the event articles, as these represent significant events in the entity's life.
For example, Table \ref{tab:benchmark_obama} shows selected events for the entity Barack Obama based on BS-ENC. In contrast to the annotations in $B_{BS-BIO}$, these events are more focused on the political happenings with major public impact. 
The benchmark $B_{BS-ENC}$ includes all relations of the specific entity to the events linked from the abstract of the Wikipedia article representing this entity.

Statistics of the entity-related information for the entities contained in the dataset $\mathcal{P}$ in the biographical sources, including in particular the number of relevant entity links and time expressions are provided in Table \ref{tab:benchmarks}. 
\ed{
Figure \ref{fig:benchmark_distribution2} illustrates the distribution of the number of relevant relations per person in the BS-BIO benchmark. Except for very few popular entities such as David Bowie and Barack Obama, the number of relevant relations is typically below $100$, with an average of $13.64$. 
}

\begin{table}[t]
	\footnotesize
	\centering
	\caption{Statistics of the dataset $\mathcal{P}$ involving $2,760$ entities of type person.}
	\label{tab:benchmarks}
	\begin{tabular}{l||r|r|r|}
		\cline{2-4}
		& \textbf{\begin{tabular}[c]{@{}l@{}}thefamous-\\ people.com\end{tabular}} & \textbf{\begin{tabular}[c]{@{}l@{}}biogra-\\ phy.com\end{tabular}} & \textbf{\begin{tabular}[c]{@{}l@{}}Wikipedia\\ Abstracts\end{tabular}} \\ \hline \hline
		\multicolumn{1}{|l||}{\textbf{Time expressions}} & 50,919 & 41,318 & 18,099 \\ \hline
		\multicolumn{1}{|l||}{\textbf{Entity links}} & 107,126 & 92,149 & 32,516 \\ \hline
	\end{tabular}
\end{table}

\colorlet{ColorLines}{black!70}
\colorlet{ColorLineMarks}{black!80}

\pgfplotsset{
    every non boxed x axis/.style={} 
}

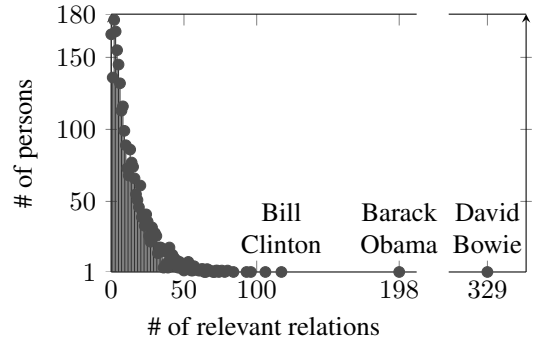
\begin{figure}
\centering
\begin{tikzpicture}
\begin{groupplot}[
    group style={
        group name=my fancy plots,
        group size=2 by 1,
        xticklabels at=edge bottom,
        horizontal sep=12pt
    },
    height=5cm,
    ymin=1,
    ymax=185
]

\nextgroupplot[
ymin=1,
xmin=0,
xmax=210,
xtick={0,50,100,198},
ytick={1,50,100,150,180},
ymax=180,
height=5cm,
width=5.6cm,
clip=false,
axis y line = left,
ylabel={\# of persons},
xlabel={\# of relevant relations}
]
\addplot+ [ycomb,mark options={ColorLines},draw=ColorLineMarks] plot coordinates
	{ (0,166) (1,136) (2,176) (3,168) (4,155) (5,145) (6,132) (7,113) (8,116) (9,99) (10,89) (11,73) (12,68) (13,86) (14,77) (15,74) (16,66) (17,55) (18,51) (19,46) (20,61) (21,39) (22,41) (23,33) (24,41) (25,36) (26,26) (27,22) (28,32) (29,27) (30,28) (31,26) (32,14) (33,18) (34,18) (35,17) (36,4) (37,15) (38,7) (39,4) (40,18) (41,10) (42,13) (43,11) (44,7) (45,9) (46,4) (47,8) (48,4) (49,3) (50,2) (51,4) (52,5) (53,8) (54,6) (55,2) (56,2) (57,5) (58,3) (60,2) (61,2) (62,3) (63,3) (64,1) (65,1) (66,3) (68,2) (70,1) (71,1) (72,2) (73,2) (74,1) (76,2) (78,1) (80,2) (84,1) (93,1) (96,1) (106,1) (117,1) (198,1) };
	\node[above] at (axis cs:117,1) {\begin{tabular}{c} Bill \\ Clinton \end{tabular}};
	\node[above] at (axis cs:198,1) {\begin{tabular}{c} Barack \\ Obama \end{tabular}};
\nextgroupplot[
ymin=1,
xmin=320,
xmax=338,
xtick=329,
ymax=180,
height=5cm,
width=2.6cm,
clip=false,
yticklabels={},
axis y line = right
]
\addplot+ [ycomb,mark options={ColorLines},draw=ColorLineMarks] plot coordinates
	{ (329,1) 	};
	\node[above] at (axis cs:329,1) {\begin{tabular}{c} David \\ Bowie \end{tabular}};
\end{groupplot}
\end{tikzpicture}
\caption{
\ed{
The number of person entities with the given number of relevant relations in the BS-BIO benchmark. The top-3 entities with the highest number of relevant relations are marked.
}
}
\label{fig:benchmark_distribution2}
\end{figure}

We generate a benchmark $B_{BS}$ for each biographical source $BS$ considered in this work.
The statistics regarding these benchmarks are presented in Table \ref{tab:instances}.

\begin{table}[]
	\centering
	\footnotesize
	\caption{Benchmark statistics: the number of entities and relevant temporal relations (temp. rel.).}
	\label{tab:instances}
	\begin{tabular}{l||r|r|r|}
		\cline{2-4}
		\textbf{} & \multicolumn{1}{r|}{\textbf{\begin{tabular}[c]{@{}r@{}}\#Per-\\ sons\end{tabular}}} & \multicolumn{1}{r|}{\textbf{\begin{tabular}[c]{@{}r@{}}\#Relevant Tem-\\ poral Relations\end{tabular}}} & \multicolumn{1}{r|}{\textbf{\begin{tabular}[c]{@{}r@{}}Avg. \# Temp.\\Rel. per Entity\end{tabular}}}  \\ \hline \hline
		\multicolumn{1}{|l||}{\textbf{$B_{BS-BIO}$}} & 2,760 &  37,638 & 13.64 \\ \hline
		\multicolumn{1}{|l||}{\textbf{$B_{BS-ENC}$}} & 2,760 &  33,106 & 12.00\\ \hline
	\end{tabular}
\end{table}

Table \ref{tab:type_distribution} provides the percentage of person types in the benchmarks. Actors and musical artists are the most frequent person types in both the training and test set.

\begin{table}[]
\centering
\caption{Percentage of top-5 entity types in the training and test set.}
\label{tab:type_distribution}
\begin{tabular}{l||r|r|}
\cline{2-3}
 & \multicolumn{1}{c|}{\textbf{Training}} & \multicolumn{1}{c|}{\textbf{Test}} \\ \hline \hline
\multicolumn{1}{|l||}{\textbf{Actor}} & 27.73\% & 28.57\% \\ \hline
\multicolumn{1}{|l||}{\textbf{Musical Artist}} & 13.32\% & 16.17\% \\ \hline
\multicolumn{1}{|l||}{\textbf{Athlete}} & 10.50\% & 6.16\% \\ \hline
\multicolumn{1}{|l||}{\textbf{Politician}} & 10.35\% & 10.44\% \\ \hline
\multicolumn{1}{|l||}{\textbf{Writer}} & 6.95\% & 11.31\% \\ \hline
\end{tabular}
\end{table}

\subsection{Classifier Setup and Timeline Statistics}
\label{sec:svm_setup}

As our binary classifier we adopted a Support Vector Machine (SVM) due to its good generalisation ability, in particular when applied to smaller datasets. We trained this classifier on the training dataset containing $1,380$ person entities, with input data normalisation, an increased weight of $3.0$ for predicting relevant instances and a linear kernel, using Weka's LibSVM implementation \cite{Weka}. From the training data, a balanced set of relevant and irrelevant instances is given to the SVM.
%

As described in Section \ref{sec:timeline_generation}, the timelines are generated by ordering the timeline entries classified as relevant chronologically by their start time. On average, each biographical timeline of the person entities in the test set contains $8.54$ entries after training the classifier on $B_{BS-BIO}$ ($B_{BS-ENC}$: $7.81$). 
Figure \ref{fig:benchmark_distribution} illustrates the number of timelines generated for the $BS-BIO$ with the specific number of entries.

\colorlet{ColorLines}{black!70}
\colorlet{ColorLineMarks}{black!80}

\pgfplotsset{
    every non boxed x axis/.style={} 
}

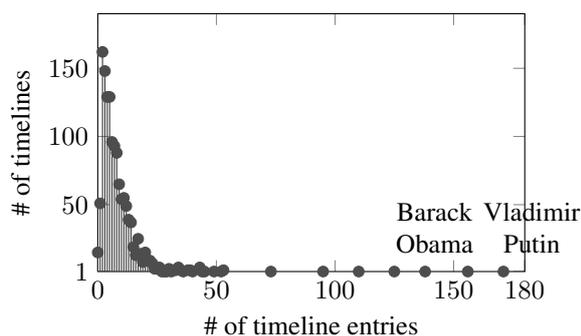
\begin{figure}
\centering
\begin{tikzpicture}
\begin{groupplot}[
    group style={
        group name=my fancy plots,
        group size=1 by 1,
        xticklabels at=edge bottom,
        horizontal sep=12pt
    },
    height=5cm,
    ymin=1,
    ymax=185
]

\nextgroupplot[
ymin=1,
xmin=0,
xmax=180,
xtick={0,50,100,150,180},
ytick={1,50,100,150},
ymax=190,
height=5cm,
width=7.2cm,
clip=false,
ylabel={\# of timelines},
xlabel={\# of timeline entries}
]
\addplot+ [ycomb,mark options={ColorLines},draw=ColorLineMarks] plot coordinates
	{ (0,15) (1,51) (2,162) (3,148) (4,129) (5,129) (6,96) (7,93) (8,88) (9,65) (10,54) (11,55) (12,49) (13,39) (14,37) (15,19) (16,13) (17,25) (18,14) (19,8) (20,15) (21,9) (22,9) (23,7) (24,3) (25,3) (26,4) (27,1) (28,1) (29,1) (30,3) (31,1) (32,2) (33,2) (34,4) (35,2) (36,1) (38,2) (39,2) (40,1) (43,4) (44,1) (45,1) (49,1) (52,1) (53,2) (73,1) (95,1) (110,1) (125,1) (138,1) (156,1) (171,1) };
	\node[above] at (axis cs:142,1) {\begin{tabular}{c} Barack \\ Obama \end{tabular}};
	\node[above] at (axis cs:183,1) {\begin{tabular}{c} Vladimir \\ Putin \end{tabular}};

\end{groupplot}
\end{tikzpicture}
\caption{
\ed{
The number of timelines with the specific number of entries generated 
for the $BS-BIO$ test set.
}
}
\label{fig:benchmark_distribution}
\end{figure}

\subsection{The TM Baseline Algorithm}
\label{sec:baseline}
We compare our proposed approach with the state-of-the-art Time Machine (TM) approach for timeline generation proposed by 
Althoff et al. \cite{Althoff:2015}. The TM approach creates events from the entity-entity relations in a knowledge graph, where one entity possesses a property with a time value. Resulting events are filtered using frequency and existence time heuristics; then a greedy algorithm selects the events that maximise a relevance score. To facilitate a fair comparison, we perform the following adjustments to implement the TM baseline:
\begin{itemize}
\item The TM approach in~\cite{Althoff:2015} was initially proposed for entity-centric knowledge graphs such as Freebase. Therefore, events in the TM terminology mean link structures in an entity-centric knowledge graph that vary with respect to their complexity. In EventKG, the events are connected to the entities directly via temporal relations. To facilitate the comparison, we adopt the TM baseline such that so-called "simple events" in the TM-terminology are generated. Such "simple events" in TM directly correspond to the temporal relations in EventKG.
\item In the original TM approach, the maximal number of temporal relations on the timeline is restricted due to the visualisation constraints; i.e. these relations are ranked by their relevance and retrieved until the visualisation constraint is met. Our goal is to provide all relevant relations, such that we do not enforce any visualisation-based constraints on the number of relations. To facilitate comparison, we retrieve an equal number of relations from the baseline and our approach. 
\item TM was initially evaluated on the Freebase dataset, and the relevance scores were computed using a search engine query log and a textual corpus. We apply all methods on the EventKG data; we use the same reference sources (i.e. Wikipedia articles) to estimate the parameters related to the global importance of entities, their occurrences and temporal relations for all baselines and approaches evaluated in this article. 
\end{itemize}

%% file: 07b_evaluation.tex
\subsection{Evaluation of the Timeline Generation}
\label{sec:evaluation}

The goals of the evaluation of the timeline generation are to assess the effectiveness of the proposed method for timeline generation and the role of the reference and biographical sources.

In particular, we assess:
\begin{itemize}
\item [G1] Quality of the generated timelines in comparison to the baseline (in a user evaluation).
\item [G2] Impact of the individual features on the timeline generation (using correlation measures). 
\item [G3] Relevance of the timeline entries with respect to the biographical source (by measuring performance of the classification model).
\item [G4] Coverage of the timeline entries with respect to the reference sources (by measuring the mean coverage of the temporal relations in the reference sources).
\end{itemize}

\subsubsection{Timeline Quality Evaluation}
\label{sec:timelinequality}

In order to evaluate the timeline quality we performed a user evaluation. We generated timelines for 60 popular entities of the types actors, athletes, musical artists, politicians and writers
for both biographical sources BS-BIO and BS-ENC.
These entities were selected from the persons in the test set described in Section \ref{sec:benchmark} based on their popularity 
(measured as the link count of the corresponding Wikipedia article). 

In each task, the user was presented with: 
(i) a task description, 
(ii) a timeline entity including its label and a Wikipedia link, and 
(iii) a pair of timelines. 
One timeline in the pair was generated by the specific configuration of our approach, 
the other timeline was generated by the TM baseline described in Section \ref{sec:baseline}. 
\ed{Both timelines were visualised as illustrated in Figure \ref{fig:timeline_example}. 
Each timeline contained all entries generated by the corresponding generation method. The user could scroll and zoom within each individual timeline. 
In the user interface, both timelines were presented simultaneously, 
one above the other, in a random order.}
We asked the users to vote for their preferred timeline in the pair. 
We provided four options: two options to vote for one of the timelines, a neutral option indicating no preference for a specific timeline, and a "don't know" option. 
We encouraged the users to research the timeline entity (e.g. using Wikipedia) 
before evaluating the timeline pair, if necessary. 

Each pair of timelines was rated by three or four users each. Then, majority voting was applied. 
In total 11 users (graduate Computer Science students) participated in the user evaluation. 
A user evaluated 42 timeline pairs on average. 
On average, the users took 69 seconds to decide between two timelines.

We compute the rater preference $RPref$ score adopted from \cite{Althoff:2015} as the fraction of votes for the particular method, based on the annotation that is most frequent among the three users per timeline entity.
The results of the user evaluation are presented in Table \ref{tab:rpref_results}.
The timelines generated by our approach with both biographical sources (BS-BIO and BS-ENC) were preferred over the baseline by the users most of the time, for all  entity types. For example, all of the $16$ timelines for politicians generated by our approach with BS-ENC were preferred over the TM timelines. In total the timelines from BS-BIO were preferred in $67.21\%$ of the cases and the BS-ENC timelines were preferred in $69.35\%$ of the cases.

\begin{table*}[]
\centering
\caption{RPRef scores from user ratings for different timeline configurations and entity types. As users could also give a neutral rating or skip a rating, the RPRef scores do not necessarily sum up to 100\%.}
\label{tab:rpref_results}
\begin{tabular}{|l||r|r||r|r|}
\hline
\textbf{Biographical Source} & \multicolumn{2}{c||}{\textbf{BS-BIO}} & \multicolumn{2}{c|}{\textbf{BS-ENC}} \\ \hline \hline
\textbf{Method} & \multicolumn{1}{c|}{\textbf{BS-BIO}} & \multicolumn{1}{c||}{\textbf{TM baseline}} & \multicolumn{1}{c|}{\textbf{BS-ENC}} & \multicolumn{1}{c|}{\textbf{TM baseline}} \\ \hline
\textbf{Actor} & 81.82\% & 9.09\% & 72.73\% & 9.09\%  \\ \hline
\textbf{Athlete} & 75.00\% & 8.33\% & 58.33\% & 25.00\% \\ \hline
\textbf{Musical Artist} & 70.00\% & 0.00\% & 50.00\% & 30.00\% \\ \hline
\textbf{Politician} & 53.33\% & 13.33\% & 100.00\% & 0.00\%  \\ \hline
\textbf{Writer} & 61.54\% & 30.77\% & 53.85\% & 25\% \\ \hline
\textbf{Total} & \textbf{67.21}\% & 13.11\% & \textbf{69.35}\% & 14.52\% \\ \hline
\end{tabular}
\end{table*}

\ed{
For BS-BIO, the mean number of ratings favouring our timeline is $1.50$ (BS-ENC: $1.58$) with a standard deviation of $0.72$ (BS-ENC: $0.97$), for the TM baseline the mean is $0.40$ (BS-ENC: $0.59$) with a standard deviation of $0.67$ (BS-ENC: $0.74$).
The results of the paired t-test confirm statistical significance of this
result for the confidence level of 99\%.
}

\subsubsection{Feature Impact}
\label{sec:feature_impact}

In total, $411$ features are utilised by the model during the timeline generation. In order to better understand the impact of the individual features on the classification task, we compute the correlation between the features and the benchmark judgements using the Pearson Correlation Coefficient ($PCC \in [-1, 1]$, with $PCC=0$ corresponding to no linear relationship), shown in Table \ref{tab:pccs}.

\begin{table*}[]
	\footnotesize
	\centering
	\caption{PCC correlation coefficient between top-5 features and the benchmark judgments, sorted by the absolute PCC values.}
	\label{tab:pccs}
\begin{tabular}{r||l|r||l|r|}
\cline{2-5}
\multicolumn{1}{l||}{} & \multicolumn{2}{c||}{\textbf{BS-BIO}} & \multicolumn{2}{c|}{\textbf{BS-ENC}} \\ \cline{1-5} 
\multicolumn{1}{|l||}{\textbf{Rank}} & \textbf{Feature} & \textbf{PCC} & \textbf{Feature} & \textbf{PCC} \\ \hline  \hline
\multicolumn{1}{|r||}{1} & TRF-PI: \textit{born} & 0.25 & TRF-PI: \textit{born} & 0.39 \\ \hline
\multicolumn{1}{|r||}{2} & TF-TP: Time provenance & 0.21 & TRF-PI: \textit{died} & 0.27 \\ \hline
\multicolumn{1}{|r||}{3} & TRF-PI: \textit{died} & 0.19 & TF-TP: Time provenance & 0.23 \\ \hline
\multicolumn{1}{|r||}{4} & TRF-MR: Relation mentions rank, EN & -0.19 & TRF-MR: Relation mentions rank, EN & -0.19 \\ \hline
\multicolumn{1}{|r||}{5} & TRF-MR: Relation mentions rank, all & -0.18 & TRF-MR: Relation mentions rank, all & -0.18 \\ \hline
\multicolumn{5}{|c|}{\textbf{…}}  \\ \hline
\multicolumn{1}{|r||}{10} & TRF-PI: \textit{spouse} & 0.13 &  TRF-MR: Relation mentions rank, RU & -0.14 \\ \hline
\multicolumn{5}{|c|}{\textbf{…}}  \\ \hline
\multicolumn{1}{|r||}{65} & TRF-PI: \textit{director} & 0.03 & TRF-PI: \textit{spouse} & 0.03 \\ \hline
\multicolumn{5}{|c|}{\textbf{…}}  \\ \hline
\multicolumn{1}{|r||}{410} & TRF-PI: \textit{cover artist} & 0.00 & TRF-PI: \textit{military rank} & 0.00 \\ \hline
\multicolumn{1}{|r||}{411} & TRF-PI: \textit{illustrator} & 0.00 & TRF-PI: \textit{draft team} & 0.00 \\ \hline

\end{tabular}
\end{table*}

For both biographical sources, the highest PCC is achieved for the property ``born'' ($PCC=0.39$ for BS-ENC, $PCC=0.25$ for BS-BIO). 
The ``died'' property and the time provenance feature TRF-TP are of similar relevance in both biographical sources, followed by the features related to relation mentions. In contrast, properties like ``cover artist'' and ``draft team'' do not correlate with the relation importance. One interesting difference between the biographical sources is the property ``spouse'' that is highly relevant in the biographical source BS-BIO, but is ranked lower in BS-ENC. Such personal happenings are often not included in Wikipedia's encyclopedic abstracts.

\subsubsection{Relevance of the Timeline Entries}
\label{sec:relevance_timeline_entries}

We evaluated the performance of the classification models for predicting the relevance of the individual temporal relations
with respect to the benchmarks presented in Section \ref{sec:benchmark}. The results of this automated evaluation using a 10-fold cross validation are presented in Table \ref{tab:accuracies}. In general, our models learned from the training set are generalisable to the test set, reaching F-measure values of $0.827$ in the case of BS-ENC and $0.738$ for BS-BIO.
Across the biographical sources, the usage of all features combined leads to the best precision and recall scores. The removal of features leads to a decrease in performance: leaving out property labels or the features based on mentions leads to the biggest performance decrease.

\begin{table*}[]
\centering
\footnotesize
\caption{Weighted precision and recall scores for both classes (relevant and irrelevant) for predicting the benchmark labels of the temporal relations using a 10-fold cross validation. Additionally, the F-measure as harmonic mean of precision and recall is reported. $\dag$ All language-dependent features except for EN are omitted.}
\label{tab:accuracies}
\begin{tabular}{ll||r|r|r||r|r|r|}
\cline{3-8}
 & \multicolumn{1}{l|}{} & \multicolumn{3}{c||}{\textbf{BS-BIO}} & \multicolumn{3}{c|}{\textbf{BS-ENC}} \\ \hline
\multicolumn{1}{|c|}{\textbf{Features}} & \multicolumn{1}{c||}{\textbf{\begin{tabular}[c]{@{}c@{}}Omitted\\ Features\end{tabular}}} & \multicolumn{1}{c|}{\textbf{\begin{tabular}[c]{@{}c@{}}Precision\end{tabular}}} & \multicolumn{1}{c|}{\textbf{\begin{tabular}[c]{@{}c@{}}Recall\end{tabular}}} & \multicolumn{1}{c||}{\textbf{\begin{tabular}[c]{@{}c@{}}F-Measure\end{tabular}}} & \multicolumn{1}{c|}{\textbf{\begin{tabular}[c]{@{}c@{}}Precision\end{tabular}}} & \multicolumn{1}{c|}{\textbf{\begin{tabular}[c]{@{}c@{}}Recall\end{tabular}}} & \multicolumn{1}{c|}{\textbf{\begin{tabular}[c]{@{}c@{}}F-Measure\end{tabular}}} \\ \hline \hline
\multicolumn{1}{|l|}{\textbf{all features}} & / & 0.796 & 0.749 & \textbf{0.738} & 0.848 & 0.829 & \textbf{0.827} \\ \hline
\multicolumn{1}{|l|}{\textbf{no property labels}} & TRF-PI & 0.753 & 0.691 & 0.671 & 0.822 & 0.802 & 0.799  \\ \hline
\multicolumn{1}{|l|}{\textbf{no mentions}} & TRF-RM & 0.769 & 0.700 & 0.679 & 0.802 & 0.734 & 0.719  \\ \hline
\multicolumn{1}{|l|}{\textbf{\begin{tabular}[c]{@{}l@{}}no temporal\\ features\end{tabular}}} & \begin{tabular}[c]{@{}l@{}}TF-TP,\\ TF-TDS,\\ TF-TDE\end{tabular} & 0.795 & 0.747 & 0.736 & 0.847 & 0.829 & 0.827 \\ \hline
\multicolumn{1}{|l|}{\textbf{English only}} & $\dag$ & 0.791 & 0.737 & 0.724 & 0.843 & 0.821 & 0.819 \\ \hline
\end{tabular}
\end{table*}

\subsubsection{Coverage of the Reference Sources}

To demonstrate the gain of integrating data from multiple reference sources into EventKG, we assess the coverage of temporal relations in the biographical sources. That means, for each person in our benchmark, we compute the percentage of benchmark relations that are found in the temporal relations of a reference source. Table \ref{tab:coverage} shows the results, measured by mean coverage per person entity. For example, $27.45\%$ of the relations extracted from BS-ENC can be mapped to a temporal relation in Wikidata. Additionally, we compute the coverage for \textit{extended} reference sources, i.e. we still only consider relations from the specific source, but use the fused information about temporal entities (i.e. existence and happening times) from EventKG.

\begin{table*}[]
\centering
\caption{Mean coverage of the temporal relations in the benchmarks per reference source and biographical source.}
\label{tab:coverage}
\begin{tabular}{l||r|r|r|r|}
\cline{2-5}
 & \multicolumn{2}{c|}{\textbf{BS-BIO}} & \multicolumn{2}{c|}{\textbf{BS-ENC}} \\ \cline{2-5} 
 & \multicolumn{1}{c|}{\textbf{Mean coverage (\%)}} & \multicolumn{1}{c|}{\textbf{\begin{tabular}[c]{@{}c@{}}Mean Coverage (\%)\\ (extended)\end{tabular}}} & \multicolumn{1}{c|}{\textbf{Mean Coverage (\%)}} & \multicolumn{1}{c|}{\textbf{\begin{tabular}[c]{@{}c@{}}Mean Coverage (\%)\\ (extended)\end{tabular}}} \\ \hline \hline
\multicolumn{1}{|l||}{\textbf{Wikidata}}  & 14.39 & 16.09 & 36.15 & 38.64\\ \hline
\multicolumn{1}{|l||}{\textbf{YAGO}}  & 11.96 & 12.34 & 37.90 & 38.40\\ \hline
\multicolumn{1}{|l||}{\textbf{Wikipedia\textsubscript{EN}}} & 0.51 & 14.56 & 0.80 & 23.65 \\ \hline
\multicolumn{1}{|l||}{\textbf{Wikipedia\textsubscript{FR}}} & 0.34 & 11.04 & 0.61 & 18.96 \\ \hline
\multicolumn{1}{|l||}{\textbf{Wikipedia\textsubscript{DE}}} & 0.16 & 0.86 & 0.40 & 16.66 \\ \hline
\multicolumn{1}{|l||}{\textbf{Wikipedia\textsubscript{PT}}} & 0.00 & 8.61 & 0.16 & 15.73 \\ \hline
\multicolumn{1}{|l||}{\textbf{Wikipedia\textsubscript{RU}}} & 0.22 & 8.68 & 0.43 & 15.41 \\ \hline
\multicolumn{1}{|l||}{\textbf{Wikipedia}} & 0.86 & 15.08 & 1.37 & 23.74 \\ \hline
\multicolumn{1}{|l||}{\textbf{DBpedia\textsubscript{EN}}} & 5.05 & 9.27 & 27.94 & 34.97 \\ \hline
\multicolumn{1}{|l||}{\textbf{DBpedia\textsubscript{FR}}} & 4.10 & 7.27 & 22.01 & 28.40 \\ \hline
\multicolumn{1}{|l||}{\textbf{DBpedia\textsubscript{DE}}} & 4.48 & 6.41 & 25.69 & 28.90 \\ \hline
\multicolumn{1}{|l||}{\textbf{DBpedia\textsubscript{PT}}} & 0.0 & 2.60 & 0.0 & 4.75 \\ \hline
\multicolumn{1}{|l||}{\textbf{DBpedia\textsubscript{RU}}} &  0.0 & 1.48 & 0.0 & 2.64 \\ \hline
\multicolumn{1}{|l||}{\textbf{DBpedia}} & 5.73 & 14.53 & 30.02 & 45.10 \\ \hline
\multicolumn{1}{|l||}{\textbf{EventKG}} & \textbf{23.29} & — & \textbf{55.09} & — \\ \hline
\end{tabular}
\end{table*}

The results show that there is a higher coverage for BS-ENC than for BS-BIO across all reference sources. This can be explained by the fact that the texts from BS-BIO are longer and less event links are provided: not only does the BS-BIO benchmark rely on named entity recognition, as this source does not contain any links, but events are also harder to recognise as they can be described in several ways (e.g. ``first inauguration of Barack Obama'' and ``Barack Obama was sworn in as the president on January 20, 2009''). In general, YAGO and Wikidata clearly outperform Wikipedia and DBpedia (as DBpedia does not contain statements with validity times). Through the integration and fusion in EventKG, the coverage increases to more than $50\%$ in BS-ENC.

%% file: 08_related.tex
\section{Related Work}
\label{sec:related}

In this section, we discuss related work in the areas of event knowledge graphs and the task of biographical timeline generation.

\subsection{Event Knowledge Graphs}
\label{sec:related-kg}

To the best of our knowledge, currently there are no dedicated knowledge graphs aggregating event-centric information and temporal relations for historical and contemporary events directly comparable to EventKG. 
The heterogeneity of data models and vocabularies for event-centric and temporal information (e.g. \cite{Shaw:2013,ROSPOCHER2016132,VanHage:2011,Guha:2011,Prasojo:2018,Yuan:2018}), the large scale of the existing knowledge graphs, in which events play only an insignificant role, and the lack of clear identification of event-centric information, makes it particularly challenging to identify, extract, fuse and efficiently analyse event-centric and temporal information and make it accessible to real-world applications in an intuitive and unified way.
Through the light-weight integration and fusion of event-centric and temporal information from different sources, EventKG enables to increase coverage and completeness of this information. 
Furthermore, existing sources lack structured information to judge event popularity and relation strength as provided by EventKG -- the characteristic that gains the key relevance given the rapidly increasing amount of event-centric and temporal data on the Web and the resulting information overload.

\textit{Data models and vocabularies for events:} 
Several data models and the corresponding vocabularies \revA{(e.g. \cite{ROSPOCHER2016132,VanHage:2011,Guha:2011,Shaw:2013, Schrodt:2012})} provide means to model events. For example, the ECKG model proposed by Rospocher et al. \cite{ROSPOCHER2016132} enables fine-grained textual annotations to model events extracted from news collections. 
\ed{
CAMEO \cite{Schrodt:2012} is a framework to model events extracted from news, in particular in the political domain.
}
The Simple Event Model (SEM) \cite{VanHage:2011}, schema.org \cite{Guha:2011} and the Linking Open Descriptions of Events (LODE) ontology \cite{Shaw:2013} provide means to describe events and interlink them with actors, times and places. 
In EventKG, we build upon SEM and extend this model to represent a wider range of temporal relations and to provide additional information regarding events. 

\textit{Extracting event-centric and temporal information:}
Most approaches for automatic knowledge graph construction and integration focus on entities and related facts rather than events. Examples include DBpedia \cite{dbpedia-swj}, Freebase \cite{Bollacker:2008}, 
YAGO \cite{Mahdisoltani:2014} and YAGO+F \cite{Demidova:2013}. 
In contrast, EventKG is focused on events and temporal relations. 
In \cite{Tran:2014}, the authors extract event information from WCEP. 
EventKG builds upon this work to include WCEP events. 
For the extraction of temporal information, there are several approaches to annotate both textual data \cite{Kuzey:2016} and relations \cite{Rula:2014,Talukdar:2012} with temporal scopes inferred from external sources. In EventKG, we rely on the temporal information already contained in the reference sources, which gives highly precise values as shown in Section \ref{sec:eval-eventkg}. Increasing the coverage for temporal annotations in case of missing values by using external resources is a potential extension for future work. 

The question of how to model temporal data is an important question as it comes to considering time expressions of different levels of granularity or with uncertainty. Examples to tackle such issues include the use of multiple potential start and end times as in the temporal slot filling task \cite{Surdeanu:2013} or adding uncertainty scores to temporal relations \cite{Chekol:2017}. The representation of this information is facilitated through existing relational models \cite{chekol2017scaling}, the Extended Date-Time Format (EDTF) \cite{EDTF} or with the Time Ontology in OWL \cite{Hobbs:2006}. The Simple Event Model adopted in this work supports a simple notion of temporal time spans, which is sufficient to represent temporal information provided by the reference sources of EventKG and is compatible with the time representation in these sources. 
Nevertheless, we see more advanced time models as a potential future extension, in particular in the context of a possible enrichment of EventKG with additional, and in particular automatically inferred, temporal information.

\textit{Extraction of events and facts from news:} Recently, the problem of building knowledge graphs and datasets directly from plain text news articles \cite{AlBadrashiny:2017, ROSPOCHER2016132, Leetaru:2013, Boschee:2015}, and extraction of named events from news \cite{Kuzey:2014, Yuan:2018} have been addressed. These approaches apply Open Information Extraction methods and develop them further to address specific challenges in the event extraction in the news domain. State-of-the-art approaches that automatically extract events from news potentially obtain noisy and unreliable results (e.g. the state-of-the-art extraction approach in \cite{ROSPOCHER2016132} reports an accuracy of only $0.551$). 
\ed{
Furthermore, such systems provide billions of events at a very high granularity level, as typically represented in news articles.
Compared to the established knowledge repositories such as DBpedia or Wikidata, such events indicate significant differences in the representation accuracy and event granularity.
In contrast, contemporary events included in EventKG originate from  high quality community curated sources such as WCEP and Wikipedia event lists and represent significant societal happenings at a different granularity and abstraction level, compared to news sources.
}

\subsection{Biographical Timeline Generation}
\label{sec:related-bio}

Existing work on timeline generation from knowledge graphs has mainly focused on the selection of relevant events or relations. The works of Althoff et al. \cite{Althoff:2015} and Tuan et al. \cite{Tuan:2011} come closest to our task definition. In \cite{Althoff:2015}, the authors create timelines for politicians, actors and athletes from the Freebase knowledge graph, adding visual and diversity constraints on the generated timelines. 
In \cite{Tuan:2011}, person timelines are generated by ranking relations extracted from Wikipedia and YAGO knowledge graphs. Similarly, in \cite{Thalhammer:2016} entity summarisation is created based on link counts, but without taking temporal data into account. In difference to our work, in both these approaches the feature weights are handcrafted and no machine learning is involved. \cite{Chisholm:2017} and \cite{Lebret:2016} aim at generating biographies in a natural language, that means to generate textual summaries for people, by mapping facts from knowledge graphs to one-sentence biographies. Both works incorporate neural models to learn text, but the biographies are limited to few facts such as birth dates and entity types.

Other approaches generate timelines for different use cases, for example to get an overview over news articles over a large time span \cite{Tran:2015, Russell:2000} or for depicting singular events such as football matches in a very fine-grained manner \cite{Alonso:2013}. For visualisation, there are approaches to transform relationship paths from knowledge graphs into sentences \cite{Althoff:2015, Voskarides:2017} and different interaction models that let a user explore the timeline \cite{Althoff:2015, Zhao:2012,Russell:2000}. In this article, we focus on the generation of timelines containing relevant temporal relations and do not limit the approach by any visual constraints. This way, the models obtained by our methods can be used in a broader range of interfaces and application scenarios.

One important subtask of the timeline generation is to judge whether a temporal relation is relevant in a certain context. This task has been addressed by other works using classification and ranking approaches. For example, to rank news articles related to a query entity, Singh et al. \cite{Singh:2016} employ a diversified ranking model based both on the aspect and temporal dimension. Approaches such as the one proposed by Setty et al. \cite{Setty:2017} impose methods to rank the importance of events, but without taking into account the specific timeline  entity. In comparison to these approaches, the task addressed in our work is more specific, as it considers the relevance of individual temporal relations to a timeline entity.

Further methods to access semantic information included in knowledge graphs in an intuitive way include question answering and spatio-temporal search applications (e.g. \cite{Zheng:2017,HoffnerWMULN17,Huang:2019, Neumaier:2018}) and interactive query construction interfaces proposed in our previous work (e.g. \cite{DemidovaZN12,Demidova:2013QC}). Application of these approaches to EventKG is an interesting direction for future research.

%% file: 09_conclusion.tex
\section{Conclusions}
\label{sec:conclusion}

In this article we presented the concept of a temporal knowledge graph
that interconnects real-world entities and events using temporal relations.
Furthermore, we presented an instantiation of the temporal knowledge graph - EventKG. EventKG is a multilingual knowledge graph that integrates and harmonises event-centric and temporal information regarding historical and contemporary events. EventKG V1.1 includes over 690 thousand event resources and over 2.3 million temporal relations.
Unique EventKG features include the light-weight integration and fusion of structured and semi-structured multilingual event representations and temporal relations in a single knowledge graph, as well as the provision of information to facilitate 
assessment of relation strength and event popularity, while providing provenance.
The light-weight integration enables to significantly increase the coverage and completeness of the included event representations, in particular with respect to time and location information.

We analysed the characteristics of the resulting knowledge graph 
and observed a significant increase in coverage compared to the reference sources. 
For example, EventKG V1.1 contains 50K more events than identified in Wikidata and more than 262K events than identified in the English DBpedia. Additionally, 360K events are extracted from semi-structured sources. The quality of this resulting dataset was confirmed in a manual evaluation. 
This evaluation indicated high precision for the event identification step (with an average precision of $96\%$), 
the time fusion step (with precision of $75\%$ for the events that had a disagreement regarding their time information in the reference sources) 
and the precision of the location fusion ($94.31\%$).

Furthermore, in this article we addressed the problem of biographical timeline generation from a temporal knowledge graph.  
In order to generate biographical timelines from a large-scale temporal knowledge graph, we proposed a method based on distant supervision.
This method uses features extracted from the temporal knowledge graph as well as a benchmark extracted from external biographical sources
to train an effective relevance model. 
Our results of a user study and an automatic evaluation 
demonstrate the effectiveness of the proposed method. Our method
significantly outperforms the baseline in the biography generation. According to the rater preference score, our method achieves $68\%$ on average, 
in contrast to the baseline that achieves only $14\%$.

We make the datasets described in this article publicly available to stimulate further research in this area.

The characteristics, statistics and evaluation results presented in this article refer to EventKG V1.1 released in March 2018. 
In February 2019, we released EventKG V2.0, briefly described in Section \ref{sec:eventkgv2}. In comparison to EventKG V1.1, EventKG V2.0 includes an increased number of events, further enhances relation fusion, provides geographical information and integrates reference sources in Italian language. 

In the future work, we plan to further extend EventKG to include additional sources. 
We would also like to explore the development of further methods and 
applications using EventKG.

%% file: 10_appendix.tex
\clearpage
\begin{appendix}

\section{Example Queries}
\label{sec:example-queries}

Here, we present example SPARQL queries to illustrate the retrieval of particular event and entity characteristics.

\subsection{Query 1: Provenance and Event Locations}

The SPARQL query in Listing \ref{lst:sparql1} uses the named graph notation to find the locations of the event ``Second inauguration of Barack Obama'' in any source. This is done using the \schema{sem:hasPlace} predicate introduced in Section \ref{sec:model}. Table \ref{tab:sparql1_results} lists the query results from EventKG V1.1. While YAGO has the United States Capitol and Washington D.C. as location, Wikidata has Washington D.C. only. There are no locations for this event found in any of the DBpedia language editions. After fusion, the union of potential locations (United States Capitol, Washington, D.C.) is reduced to the United States Capitol only, which is located in Washington D.C\footnote{This information could be inferred using \schema{so:con\-tained\-In\-Place\textsuperscript{*}}.}. Fused locations are placed within EventKG's named graph.

\begin{table}[h]
\centering
\caption{Locations of the first inauguration of Barack Obama in EventKG.}
\label{tab:sparql1_results}
\begin{tabular}{|l|l|}
\hline
\multicolumn{1}{|c|}{\textbf{\schema{\textbf{?location}}}} & \multicolumn{1}{|c|}{\textbf{\schema{?named\_graph}}} \\ \hline \hline
dbr:United\_States\_Capitol & eventKG-g:event\_kg \\ \hline
dbr:Washington,\_D.C. & eventKG-g:wikidata \\ \hline
dbr:United\_States\_Capitol & eventKG-g:yago \\ \hline
dbr:Washington,\_D.C. & eventKG-g:yago \\ \hline
\end{tabular}
\end{table}

\newpage

\subsection{Query 2: Important Events of an Entity}

The second query shown in Listing \ref{lst:sparql2} employs the relation strength information contained in EventKG. It returns a list of events connected to Barack Obama, sorted by the number of common mentions (\schema{eventKG-s:mentions}) with Barack Obama in the English Wikipedia (\schema{GRAPH eventKG-g:wikipedia\_en}). Additionally, if there is an event start date available, this is returned as well, using the named EventKG graph to retrieve the fused date. The results from EventKG V1.1 in Table \ref{tab:example_query_results_obama_events} reveal that the United States presidential election of 2008 is the event mentioned most often together with Barack Obama. 

\begin{table}[h]
\scriptsize
\centering
\caption{Events that are most often mentioned together with Barack Obama.}
\label{tab:example_query_results_obama_events}
\begin{tabular}{|l|l|l|}
\hline
\multicolumn{1}{|c|}{\textbf{\schema{?event}}} & \multicolumn{1}{|c|}{\textbf{\schema{?cnt}}} & \multicolumn{1}{|c|}{\textbf{\schema{?startDate}}} \\ \hline \hline
dbr:United\_States\_presidential\_election,\_2008 & 719 & 2008-11-04 \\ \hline
\begin{tabular}[c]{@{}l@{}}dbr:United\_States\_presidential\_election \\ \_in\_New\_Jersey,\_2012\end{tabular} & 530 & 2012-11-06 \\ \hline
\begin{tabular}[c]{@{}l@{}}dbr:United\_States\_presidential\_election \\ \_in\_New\_Jersey,\_2008\end{tabular} & 522 & 2008-11-04 \\ \hline
\multicolumn{3}{|c|}{⋮} \\ \hline
dbr:First\_inauguration\_of\_Barack\_Obama & 68 & 2009-01-20 \\ \hline
\end{tabular}
\end{table}

\clearpage

\begin{lstlisting}[float=*,captionpos=b, caption={SPARQL query for retrieving the locations of the first inauguration of Barack Obama using \schema{sem:hasPlace}, together with their named graph for provenance information.}, label=lst:sparql1,   basicstyle=\ttfamily,frame=no]
SELECT ?location ?named_graph

WHERE {
  ?event owl:sameAs dbr:First_inauguration_of_Barack_Obama .
  
  GRAPH ?named_graph {
    ?event sem:hasPlace ?loc
  } .
  
  GRAPH eventKG-g:dbpedia_en {
    ?loc owl:sameAs ?location .
  }
}
ORDER BY ?named_graph
\end{lstlisting}

\begin{lstlisting}[float=*,captionpos=b, caption=SPARQL query for retrieving the events that are most often mentioned together with Barack Obama. Instances of \schema{eventKG-s:Relation} are searched who are connected to Barack Obama as their subject and an instance of \schema{sem:Event} as their object., label=lst:sparql2,
   basicstyle=\ttfamily,frame=no]
SELECT ?event ?cnt ?startDate

WHERE {
  ?obama owl:sameAs dbr:Barack_Obama .
  ?relation rdf:subject ?obama .
  ?relation rdf:object ?eventEKG  .
  
  GRAPH eventKG-g:wikipedia_en {
    ?relation eventKG-s:mentions ?cnt .
  }
  
  ?eventEKG rdf:type sem:Event .
  
  GRAPH eventKG-g:dbpedia_en {
    ?eventEKG owl:sameAs ?event
  } .
  
  OPTIONAL {
    GRAPH eventKG-g:event_kg {
      ?eventEKG sem:hasBeginTimeStamp ?startDate 
    }
  } .
}
ORDER BY DESC(?cnt)
\end{lstlisting}

\end{appendix}